\pdfoutput=1
\documentclass[11pt]{article}

\usepackage{acl}

\usepackage{times}
\usepackage{latexsym}

\usepackage[T1]{fontenc}
\usepackage[utf8]{inputenc}

\usepackage{microtype}

\usepackage{inconsolata}

\usepackage{graphicx}

\usepackage{amsmath}
\usepackage{subcaption}

\usepackage{xurl}
\usepackage{hyperref}

\usepackage{fancyhdr}

\usepackage{amssymb}

\title{Interpretable Stylistic Variation in Human and LLM Writing Across Genres, Models, and Decoding Strategies}

\author{
\bfseries
\textbf{Swati Rallapalli$^{1}$, Shannon K. Gallagher$^{2}$, Ronald Yurko$^{2}$, Tyler Brooks$^{1}$} \\
\textbf{Chuck Loughin$^{1}$, Michele Sezgin$^{2}$, Violet Turri$^{1}$} \\
$^{1}$Software Engineering Institute, AI Division, Carnegie Mellon University \\
$^{2}$Department of Statistics \& Data Science, Carnegie Mellon University
}

\begin{document}
\pagestyle{fancy}
\fancyhead{}
\renewcommand{\headrulewidth}{0pt} % no line in header area
\fancyfoot{} % clear all footer fields
\fancyfoot[LE,RO]{\thepage}    
\fancyfoot[RE,LO]{[Distribution Statement A] Approved for public release and unlimited distribution.}

\fancypagestyle{empty}{
\fancyhead{}
\renewcommand{\headrulewidth}{0pt} % no line in header area
\fancyfoot{} % clear all footer fields
\fancyfoot[LE,RO]{\thepage}    
\fancyfoot[RE,LO]{[Distribution Statement A] Approved for public release and unlimited distribution.}
}

\maketitle
\begin{abstract}
 Large Language Models (LLMs) are now capable of generating highly fluent, human-like text. They enable many applications, but also raise concerns such as large scale spam, phishing, or academic misuse. While much work has focused on detecting LLM-generated text, only limited work has gone into understanding the stylistic differences between human-written and machine-generated text. In this work, we perform a large scale analysis of stylistic variation across human-written text and outputs from 11 LLMs spanning 8 different genres and 4 decoding strategies using Douglas Biber's set of lexicogrammatical and functional features. Our findings reveal insights that can guide intentional LLM usage. First, key linguistic differentiators of LLM-generated text seem robust to generation conditions (e.g., prompt settings to nudge them to generate human-like text, or availability of human-written text to continue the style); second, genre exerts a stronger influence on stylistic features than the source itself; third, chat variants of the models generally appear to be clustered together in stylistic space, and finally, model has a larger effect on the style than decoding strategy, with some exceptions. These results highlight the relative importance of model and genre over prompting and decoding strategies in shaping the stylistic behavior of machine-generated text.
\end{abstract}

\section{Introduction}
\label{sec:intro}

As Large Language Models (LLMs) have seen significant advancement in recent times, they are demonstrating fluency and proficiency across a wide range of tasks~\cite{brown2020language, openai2023gpt4}. As a result, the distinction between human-written text (HWT) and machine-generated text (MGT) is becoming increasingly blurred, making it challenging to differentiate between the two. While this can benefit applications such as conversational agents and assistive writing tools, it also raises concerns regarding traceability and misuse. For instance, misuse such as large scale spam, phishing attacks or academic plagiarism~\cite{brown2020language} is of concern. Furthermore, prior work~\cite{Spitale_2023} shows that GPT-3-generated misinformation can be as hard or harder to recognize than human-generated misinformation. 

Therefore, a lot of work has gone into classification of human-written text and machine-generated text. Existing approaches include supervised classifiers~\cite{openai_detector}, methods based on language model probability signals~\cite{gltr,detectgpt}, and alternative techniques such as watermarking~\cite{watermark01}. A comprehensive survey of these methods is provided by~\cite{raid_analysis_emnlp2025}. While it is interesting to understand if we are able to successfully distinguish between LLM-generated and human-written text and the related nuances (e.g., how are the results related to the complexity of the model), it is also important to understand the linguistic features that drive the distinction as they provide deeper insights into how LLMs learn and generate language and the extent to which they adhere to or diverge from the stylistic patterns observed in human language.

The analysis of stylistic patterns is also of growing interest to users. For instance, recent observations have highlighted characteristic stylistic tendencies in ChatGPT-generated text, such as the frequent use of em dashes, which has attracted considerable attention~\footnote{\url{https://www.nytimes.com/2025/09/18/magazine/chatgpt-dash-hyphen-writing-communication.html}}. Some users seek to understand the underlying reasons for such patterns, while others aim to avoid them due to concerns about their association with machine-generated text. All in all this suggests the need for systematic understanding of stylistic features of machine-generated language. However, there is limited but growing body of work that explicitly characterizes the linguistic differences between human-written and machine-generated text. 

One important paper in this direction is \citet{llm_human_biber_yurko}. They construct two parallel corpora of human-written and LLM-generated texts from common prompts using several variants of Llama 3 and GPT-4o. With these corpora, they analyze Douglas Biber's set of lexical, grammatical, and rhetorical features \cite{Biber:1988, Biber:1995, Biber:2009} to identify that LLMs struggle to match human stylistic variation. They observe that systematic differences in the Biber features persist across model sizes, with larger differences for instruction-tuned models compared to base models. 

Motivated by this line of work, we analyze the large-scale RAID dataset~\cite{raid}, which contains human-written and LLM-generated text across 11 models, 8 genres and 4 decoding strategies using the Biber features. The goal is 
to understand LLM-generated text under different prompting strategies, across different genres and decoding methods -- factors not explored in prior work. Also, unlike the Reinhart study, the RAID dataset does not provide models with access to initial human-written context, or explicitly prompt them to mimic human writing styles. 

Ours findings are as follows: (i) key linguistic characteristics of machine-generated text remain robust across generation conditions and prompt settings; in particular, LLM outputs tend to be more information dense and noun-heavy, especially for instruction-tuned chat models, consistent with prior observations despite differences in data generation settings; (ii) the genre plays a stronger role than the source of text (i.e., human or a specific LLM), with texts from different sources but the same genre exhibiting more similar linguistic features than texts from the same source across genres, suggesting that LLMs adapt their style of writing for different genres (e.g., news versus reviews), even without explicit prompting; (iii) instruction-tuned models are frequently clustered separately from both base models and human-written text, aligning with prior findings; and (iv) model choice has a larger effect on stylistic variation than decoding strategy (i.e., algorithm to pick next token from model output space), with some exceptions. These results highlight the relative importance of model and genre over prompting and decoding strategies in shaping the stylistic behavior of machine-generated text, and provide guidance to users when selecting among different LLM configurations.

\section{Related Work}
\label{sec:related} 

In this Section, we focus on the few other works that analyze linguistic features of machine-generated text.

A related work is \cite{raid_analysis_emnlp2025}. They analyze the RAID dataset for different linguistic structures. Unlike our analysis, they do not use the 67 Biber features and the analysis goals are also complementary to ours. Using a set of linguistic features related to morphology, syntax, semantics and other features like text length, they find that human-written text tends to exhibit simpler syntactic structures and more diverse semantic content. They further measure variability across categories and also newer versus older models using the above linguistic features as well as style embeddings~\cite{patel-etal-2025-styledistance} and the results are mixed based on the features used and therefore need further investigation. Prior work by the same authors~\cite{zanotto1_arxiv} similarly highlights greater variability in linguistic features for human-written text compared to LLM-generated text. Additionally,~\cite{sandler2024} compares human and ChatGPT-generated dialogues using Linguistic Inquiry and Word Count (LIWC) features, finding that human conversations exhibit greater variability and authenticity. Collectively, these studies emphasize variability in linguistic features, which is not the primary focus of our analysis. 

Other recent work has examined linguistic differences between human-written and LLM-generated text, identifying systematic variations in stylistic and structural patterns~\cite{reddit_analysis, news_analysis} in specific settings. 
A recent survey~\cite{linguistic_analysis_survey} reviews existing work on the linguistic characteristics of machine-generated text and highlights the need for more large-scale and systematic analyses. In this work, we conduct a more comprehensive stylistic analysis of machine-generated text across diverse models, genres, and generation settings. 
\section{Approach}
\label{sec:approach}
In this Section we overview the dataset and the features used in our analysis.

\subsection{Dataset}
We use the RAID dataset~\cite{raid}, a large-scale benchmark generated for machine-generated text detection. RAID comprises over 6 million text samples spanning 12 sources, including human-written text and outputs from 11 large language models (LLMs): GPT-2 XL, GPT-3, GPT-4, ChatGPT, Mistral-7B, Mistral-7B (Chat), MPT-30B, MPT-30B (Chat), LLaMA 2 70B (Chat), Cohere, and Cohere (Chat). The dataset covers 8 genres: Abstracts, Books, News, Poetry, Recipes, Reddit, Reviews, Wikipedia and includes 11 adversarial attacks. Text is generated under 4 decoding configurations, combining greedy decoding and random sampling, each with and without repetition penalties where applicable.

The dataset contains 13,371 human written documents, across the above 8 genres (roughly 2,000 documents for each category). For each human-written document one output is generated for each of the 11 models, 4 decoding strategies and 11 adversarial attacks. Generation is performed in a zero-shot manner using prompt templates based on the model type. “Chat” models fine-tuned on dialogue are prompted using chat-style instructions (e.g., \texttt{Write the abstract for the academic paper titled \{title\}}), while continuation-based “Non-Chat” models are prompted with prefix-style templates (e.g., \texttt{The following is the full text of the abstract for a research paper titled \{title\} from arxiv.org:}). 

The prompts are otherwise similar across settings, with the \texttt{{title}} field replaced with the title of the corresponding human-written document.
Importantly, no explicit constraints are imposed on properties such as style or length, allowing the generations to more closely reflect realistic scenarios.

Most prior works use this dataset for detection of machine-generated text. We, on the other-hand, use it to study the lexical and stylistic properties of human-written versus machine-generated text. We use a subset of this dataset which excludes texts generated using adversarial attacks in our analysis. The goal of the selected subset is to reflect common generation settings. Therefore, in the end our analysis dataset contains 467,985 number of texts. 

The dataset used for the analysis in~\cite{llm_human_biber_yurko} used a different set-up. First, it used a smaller-set of 6 models, namely, multiple versions of GPT-4o~\cite{openai2024gpt4o, openai2024gpt4omini} and Meta Llama3~\cite{meta2024llama3}. Second, the prompt settings are different. Instruction-tuned models are prompted with: \texttt{“In the same style, tone, and diction of the following text, complete the next 500 words, generate exactly 500 words, and note that the text does not necessarily end after the generated words:”}, followed by an initial chunk of human-written text. The base (non-instruction-tuned) Llama 3 variants are provided only with the initial text chunk as a prefix. In either case, the models generate approximately 500 words of output, extending the given human-written text which was then compared against the corresponding second chunk of the original human-written document. 

The differences in dataset construction lead to the following distinctions in our analysis. First, our setting captures a different range of stylistic variation: RAID generations are produced without explicit prompts to mimic human style, unlike~\cite{llm_human_biber_yurko}, where models are guided to closely match human-written text and are provided with an initial text chunk. Second, we analyze genre
level differences among the texts generated by different sources. Third, we analyze the stylistic differences stemming from the different decoding styles available in the RAID dataset.

\subsection{Features}
As in~\cite{llm_human_biber_yurko} paper, we use 67 Biber features~\cite{biber_org}, extracted using the \texttt{pseudobibeR}~\cite{biber_pkg} package. These features capture a broad range of grammatical and structural properties of text, including tense and aspect markers, clause types, nominalizations, and other syntactic constructions. Biber features are designed to quantify stylistic variation by characterizing how language is used, such as differences in information density or narrative style rather than what is being said (i.e., topic or semantic content). A detailed description of the feature set can be found in~\cite{biber_pkg} and Appendix Table 2 of~\cite{llm_human_biber_yurko}. The rate of occurrence of each feature is counted per 1000 words of text. These rates serve as features for our analysis.  

\begin{figure}[ht]
    \vspace{-2mm}
    \begin{center}
    \includegraphics[width=0.48\textwidth]{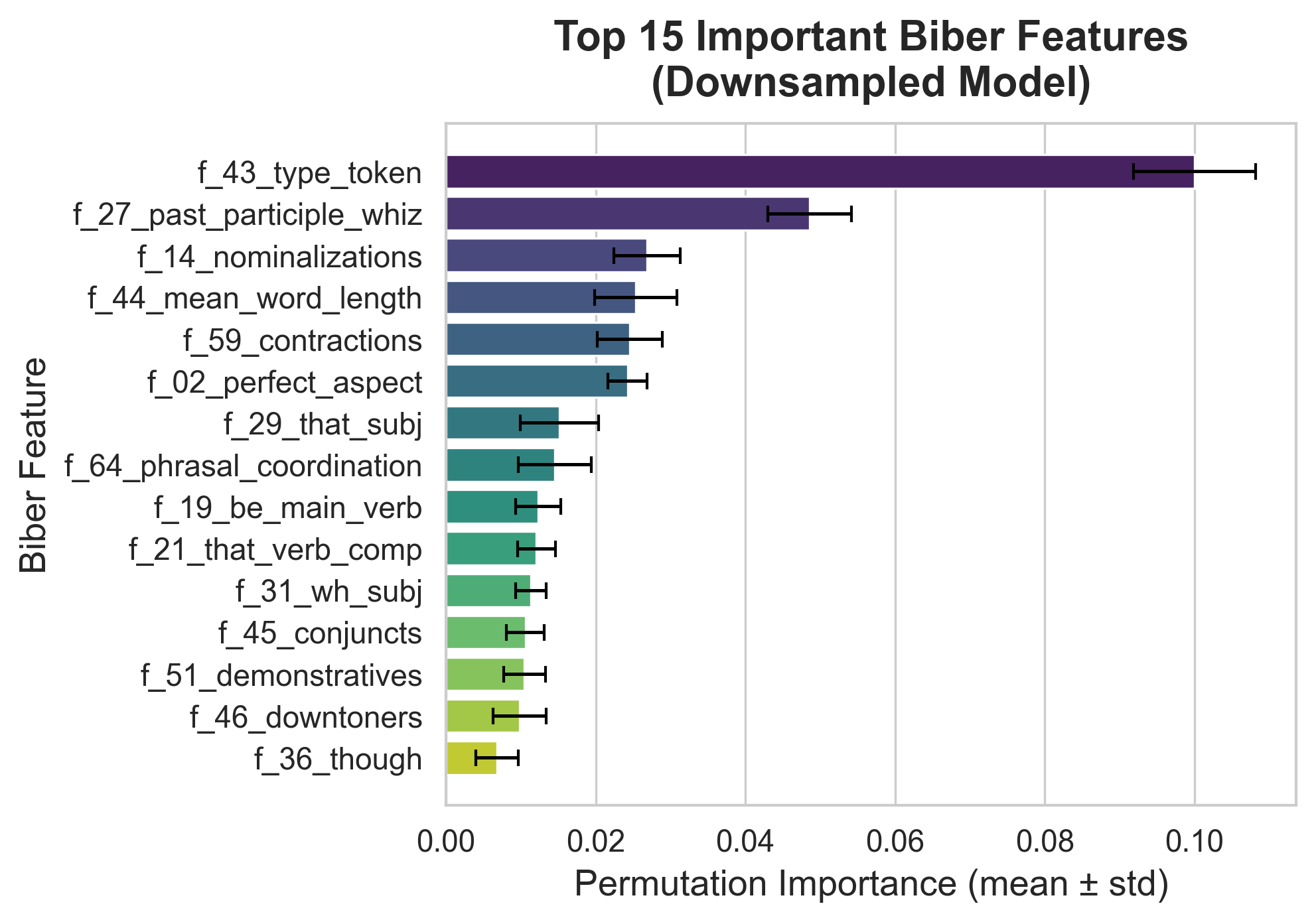}
    \caption{Top 15 most important Biber features for distinguishing human and LLM-generated text based on Random Forest classifier on down-sampled model.}
    \label{fig:top15_rf_features}
    \end{center}
    \vspace{-2mm}
\end{figure}

\begin{figure}[ht]
    \vspace{-2mm}
    \begin{center}
    \includegraphics[width=0.48\textwidth]{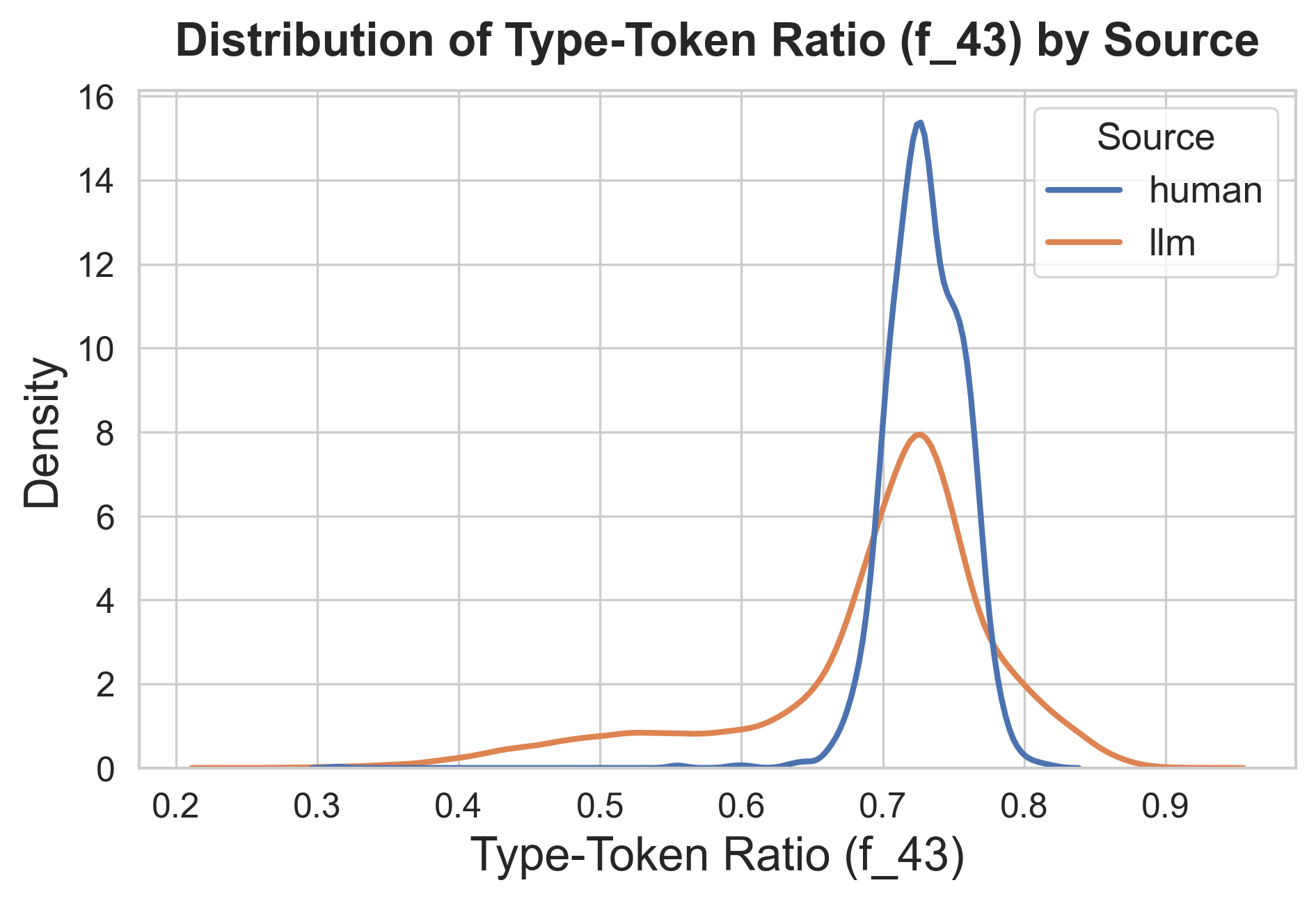}
    \caption{Feature Distribution for \texttt{f\_43\_type\_token}}
    \label{fig:distribution}
    \end{center}
    \vspace{-3mm}
\end{figure}

\begin{figure}[ht]
    \vspace{-2mm}
    \begin{center}
    \includegraphics[width=0.48\textwidth]{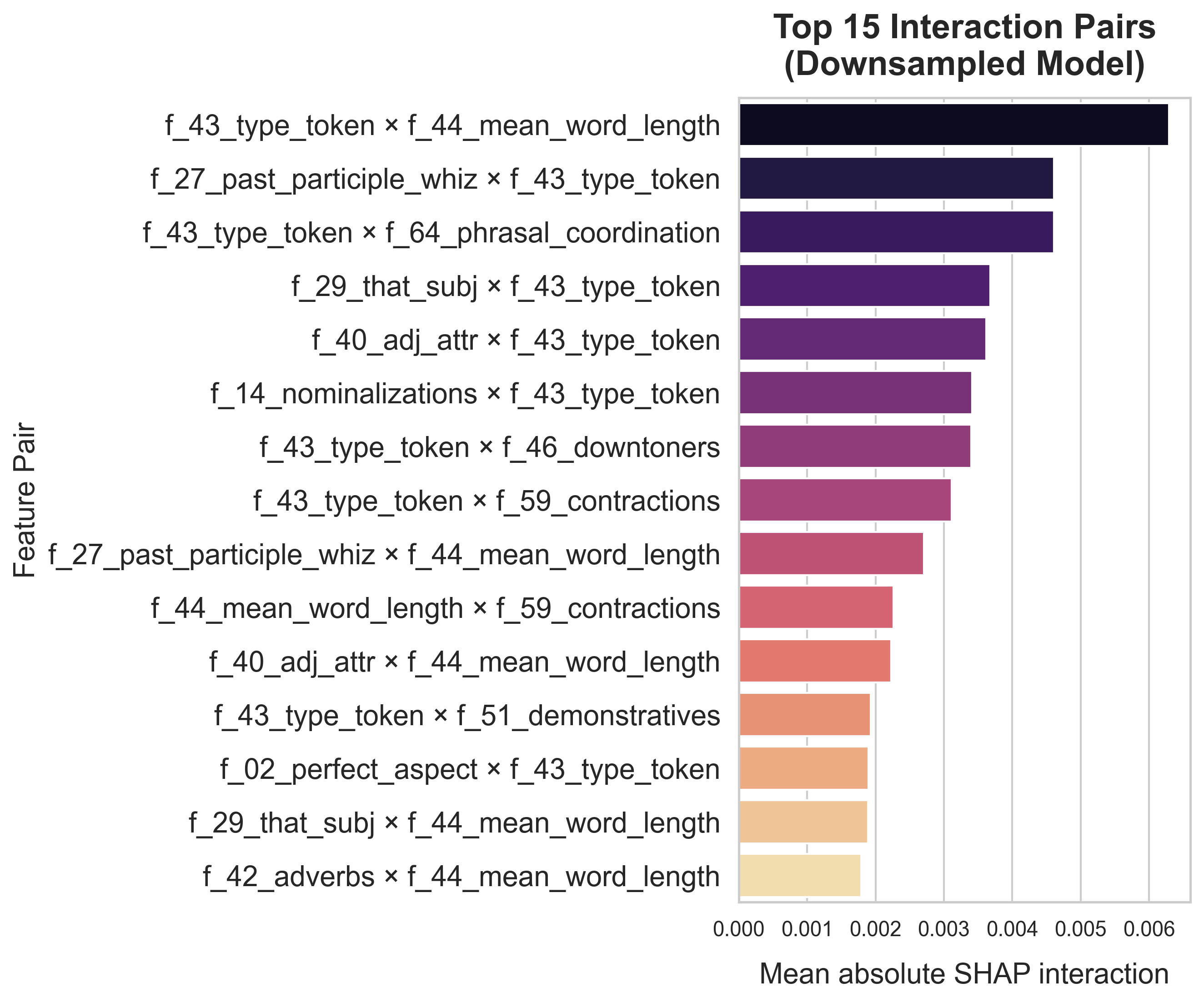}
    \caption{Top 15 SHAP interaction pairs based on Random Forest classifier on down-sampled model.}
    \label{fig:top15_shap_interaction}
    \end{center}
    \vspace{-3mm}
\end{figure}

\subsubsection{Important Biber features based on classification}
\label{sssec:classification}
While we look at 67 Biber features in our analysis, it is useful in some cases to zoom in on a few. Moreover, a dimension to understand is which of these features are actually helpful in discriminating between human-written text (HWT) and machine-generated text (MGT). The goal of this work is not the classification itself, but understanding the most discriminative features. 

To this end, we train a Random Forest Classifier (with $200$ trees) using Biber features on the RAID data. The dataset is inherently imbalanced, as each human-written document is paired with outputs from 11 LLMs. Therefore, we train two models: (i) a base model using the full imbalanced data, and (ii) a down-sampled model in which the majority class (MGT) is reduced. We use a $60\%/20\%/20\%$ train/validation/test split. For the down-sampled model, we tune the class ratio on the validation set using F1 score for the human class. A ratio of 4:1 (MGT:HWT) yields the best performance. We then train both models on combined train and validation data and evaluate on the held-out test set.

On the test set, the down-sampled model achieved an F1 score of 0.67 for the human class and 0.99 for the LLM class, compared to 0.21 and 0.98, respectively, for the base model. However, the AUC is very close for both the models: 0.9775 for down-sampled model and 0.9755 for base-model, that is the discriminative ability of both models is comparable with a slight advantage with the down-sampled model. The corresponding ROC curves are shown in Appendix Figure~\ref{fig:roc}. 
We use permutation importance from the down-sampled model to identify the top 15 most discriminative features, which are the focus of our analysis. % in this paper. 

Figure~\ref{fig:top15_rf_features} presents these top 15 features. We further examine them to understand the factors driving their discriminative power. Notably, the most important feature \texttt{f43\_type\_token} is a measure of lexical diversity or vocabulary richness, defined as the ratio of unique words (types) to the total number of words (tokens) in a given text. As we will see in Section~\ref{sec:results}, it is neither overused nor underused by LLMs. Then to better understand its importance, we analyze its distribution in Figure~\ref{fig:distribution}. This shows that for humans, it is narrow and highly predictable around the mean value. The distribution for all the features is in the Appendix Figures~\ref{fig:biber_dist_part1} and~\ref{fig:biber_dist_part2}. 

Also, in Section \ref{sec:results} we look at features that are overused or underused by LLMs, and we notice that 4 of the top 5 overused features are among the top 15 important features. There is also some representation from the under-used features in the top 15 most discriminatory feature set. The rest are likely due to the difference in distribution as we see in Type Token (feature 43) or their role in interactions with other features.

To investigate this further, we look at the top interacting features using SHAP interaction values\footnote{\href{https://shap.readthedocs.io/en/latest/generated/shap.TreeExplainer.html}{SHAP TreeExplainer documentation}} in Figure~\ref{fig:top15_shap_interaction}.
This allows us to better understand why certain features are important and to identify combinations of stylistic features that jointly contribute to distinguishing human-written and machine-generated text. 

We observe that features involved in the strongest interaction pairs largely overlap with the most important features identified earlier via permutation importance. This is likely because the features participating in strong interactions are present at multiple splits in the tree and therefore govern the feature importance. Note that \texttt{f\_43\_type\_token} appears in a large fraction of the highest-ranked SHAP interaction pairs, suggesting that it modulates the predictive effect of multiple other linguistic features and also explains why it is an important feature for classification.

\section{Results}
\label{sec:results}
%This Section presents the results of our analysis. 

\subsection{Feature Overuse/Underuse by LLMs}

We analyze overuse and underuse of Biber features in LLM-generated text relative to human-written text, both in aggregate across models and at the level of individual LLMs. %We look at feature usage ratios as explained below. 

\subsubsection{Aggregate over all LLMs} 
\label{sssec:aggregate_overuse_underuse}
We first analyze aggregate feature usage by computing ratios of feature usage across all LLMs relative to human-written text. For each feature, we compute the mean and standard error across all LLM-generated samples and normalize these by the corresponding mean feature usage in human-written text. We then focus on the: (i) top 5 overused and underused features, and (ii) visualize the ratios for the remaining features.

\paragraph{Top overused/underused features:}

We identify the five most overused and underused Biber features in LLM-generated text, considering all features (not limited to those in Figure~\ref{fig:top15_rf_features}).

\noindent \emph{(1) Top overused features in LLM-generated text:}
\begin{enumerate}
    \item \texttt{f\_29\_that\_subj}: \textit{That}-clauses as subject (e.g., the dog \textit{that bit me})
    \item \texttt{f\_59\_contractions}: Contractions (e.g., \textit{can't}, \textit{won't})
    \item \texttt{f\_27\_past\_participle\_whiz}: Past participial postnominal clauses (e.g., The solution \textit{produced by this process})
    \item \texttt{f\_14\_nominalizations}: Nominalizations (noun forms derived from verbs or adjectives)
    \item \texttt{f\_34\_sentence\_relatives}: Sentence relatives (e.g., Bob likes fried mangoes, \textit{which is the most disgusting thing I've ever heard of}.) 
\end{enumerate}

Some of these features contribute to a more information dense, noun-heavy style, particularly nominalizations and past participal postnominal clauses while others reflect more elaborative or conversational nature of LLM-generated text.

\noindent \emph{(2) Top underused features in LLM-generated text:}
\begin{enumerate}
    \item \texttt{f\_32\_wh\_obj}: \textit{Wh}-relative clauses functioning as objects (e.g., the man \textit{who Sally likes})
    \item \texttt{f\_33\_pied\_piping}: Pied-piping relative clauses: relative clauses moved in the sentence by 'wh-' questions (e.g., the manner \textit{in which he was told})
    \item \texttt{f\_36\_though}: Concessive subordinators (e.g., \textit{though}, \textit{although})
    \item \texttt{f\_66\_neg\_synthetic}: Synthetic negation (e.g., \textit{No answer is good enough})
    \item \texttt{f\_50\_discourse\_particles}: Discourse particles in sentence-initial position (e.g., \textit{well}, \textit{now}, \textit{anyway})
\end{enumerate}

These features generally reflect complex style of writing and are relatively infrequent even in human text (they have a mean human frequency between $0.2-1.2$, whereas for the top overused features it is between $1-14$), suggesting that LLMs further suppress already rare constructions.

\noindent \emph{(3) Observations:} Note that most of the top 5 overused features are among the important features for classification (Figure~\ref{fig:top15_rf_features}). Only one of the top 5 underused features is among top 15 important features. This likely represents the fact that the underused features are under-represented in both human-written and machine-generated text, limiting their discriminatory ability.

\paragraph{Visualize ratios for other features:} We look at the ratio of mean feature usage of LLMs to Humans for top 15 most discriminatory features in Figure \ref{fig:ratio_plot_imp} (and all the features in Appendix Figure \ref{fig:ratio_plot_all}). Note the log scale for plotting (ratio is not log, just the visualization is in log scale). Some important features are overused, while some are underused. Some are nearly similarly used as humans, as we see in Figure~\ref{fig:distribution}, they may still be important in classifying human-written versus machine-generated text, due to distribution differences and their role in feature interactions. 

Another aspect worth noting is that even though the datasets used for the analysis are very different and models analyzed have only some overlaps, \cite{llm_human_biber_yurko} also found that instruction-tuned models favor present participial clauses, ‘that’ clauses as subjects, nominalization, and phrasal co-ordination, which are generally markers of more information dense, noun-heavy style of writing. Most of these features are top overused features even in our aggregate analysis (in Section~\ref{sssec:individual_overuse_underuse} we will also see that overuse is more pronounced in the instruction-tuned "Chat" models). This is interesting because of the differences in the experimental setup: in \cite{llm_human_biber_yurko}, models were conditioned on the first 500 tokens of human-written text and explicitly prompted to mimic human style, whereas neither condition holds for the RAID dataset used in our analysis.

\begin{figure}[ht]
    \vspace{-2mm}
    \begin{center}
    \includegraphics[width=0.48\textwidth]{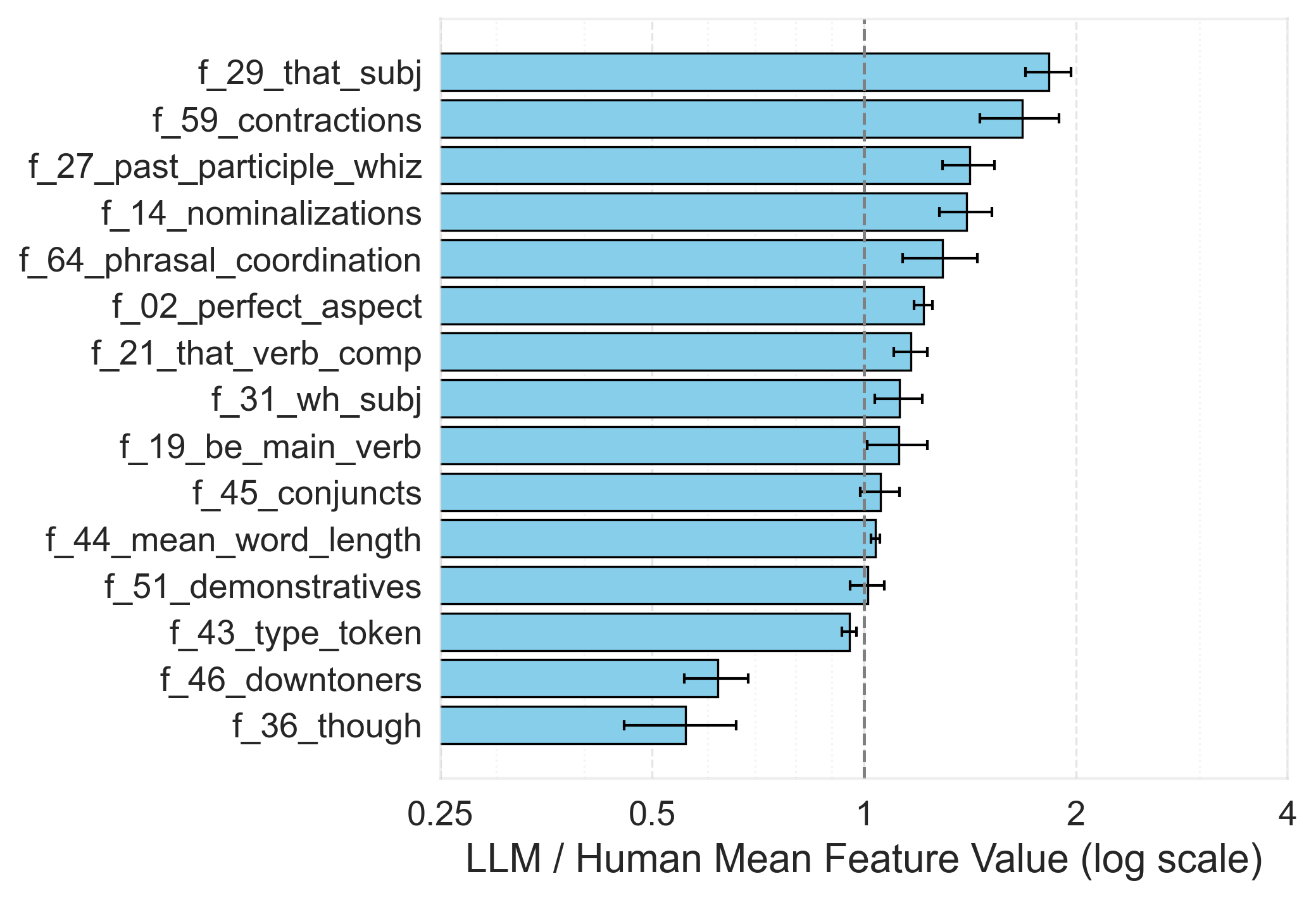}
    \caption{Ratio of Feature Usage LLM Vs. Human for Selected Features.}
    \label{fig:ratio_plot_imp}
    \end{center}
    \vspace{-4mm}
\end{figure}

\subsubsection{Individual LLMs}
\label{sssec:individual_overuse_underuse}
\begin{figure*}[ht]
    \centering
    \includegraphics[width=0.85\textwidth]{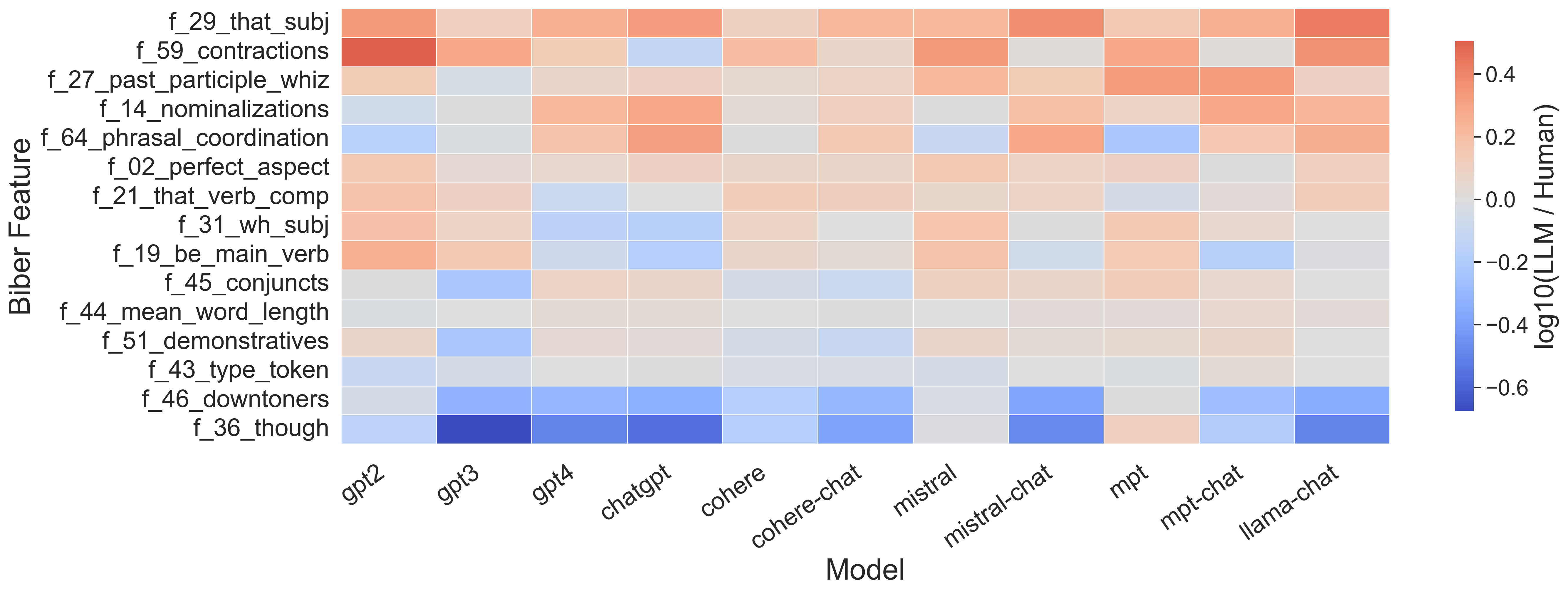}
    \caption{Heat map of Log10 Biber Feature Ratios (LLM / Human) for Selected Important Features}
    \label{fig:heat_map_imp}
\end{figure*}

We look at the ratio of feature usage of individual LLMs to humans for top 15 selected features in Figure~\ref{fig:heat_map_imp}. We plot the log of ratio for visual clarity. This reveals some interesting trends. First, many features seem to be showing similar trends across LLM models. For instance \texttt{f\_29\_that\_subject} is overused significantly by all models, \texttt{f\_46\_downtoners}
is underused by all. Some features show a mixed trend for e.g., \texttt{f\_31\_wh\_subj}, \texttt{f\_19\_be\_main\_verb}. This is also the trend we see in the 
corresponding figure for all features is in the Appendix~\ref{fig:heat_map_all}. Second, to re-iterate what we mentioned in Section \ref{sssec:aggregate_overuse_underuse}, despite differences in experiment settings, we observe trends similar to \cite{llm_human_biber_yurko}. Specifically, they found that instruction-tuned models tend to favor present participial clauses, ‘that’ clauses as subjects, nominalization, and phrasal co-ordination, which are characteristics of more information dense, noun-heavy style of writing. These features are among the top overused features even in our previous aggregate analysis. Here, additionally, we would like to highlight that the overuse is more pronounced in the instruction-tuned "Chat" models. Finally, we notice certain models are closer to human writing than others, for e.g., the Cohere model exhibits feature usage ratios that are consistently closer to 1.

\subsection{Visualization of Features Across Sources and Genres}
\begin{figure*}[ht]
    \centering
    \includegraphics[width=0.85\textwidth]{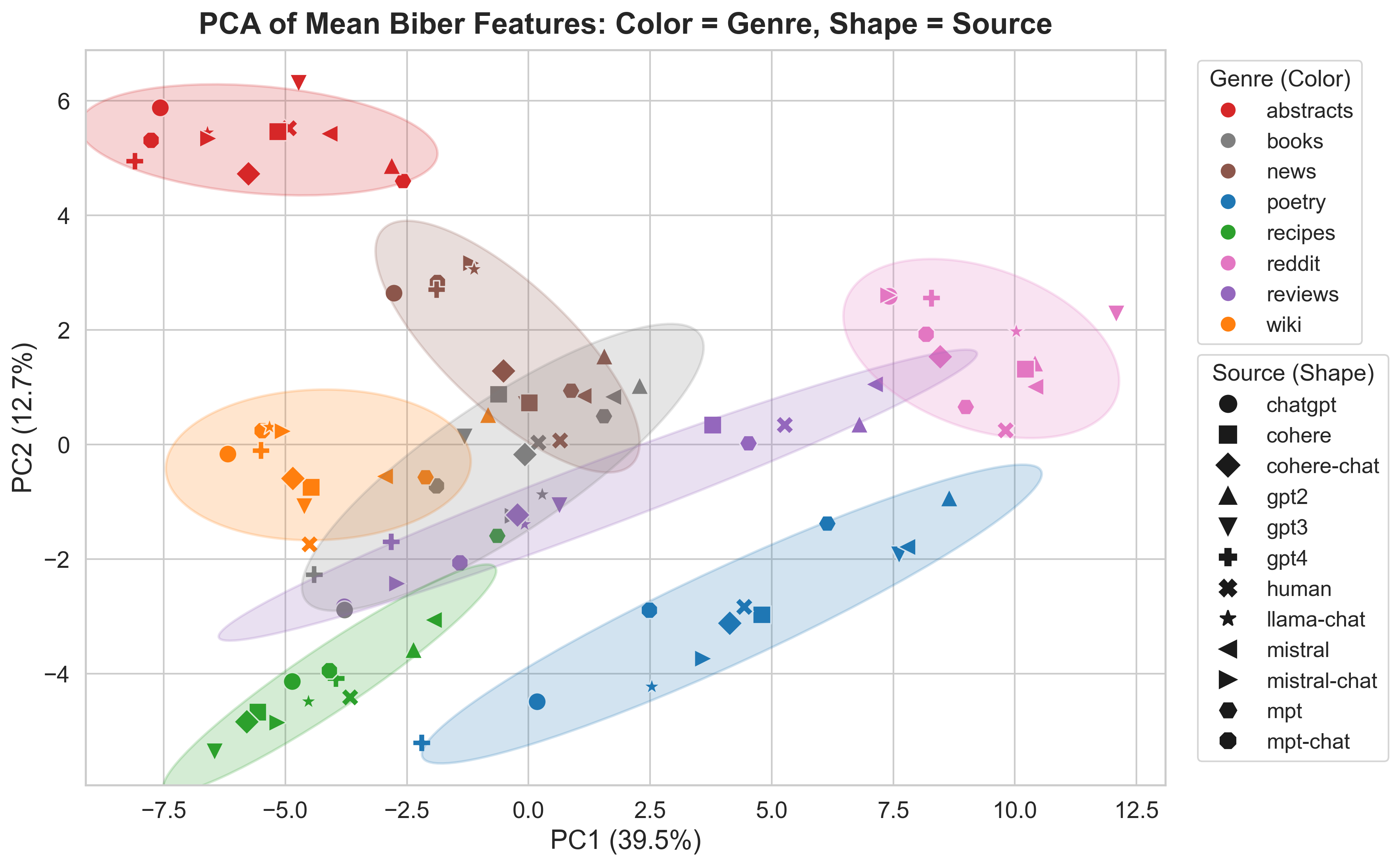}
    \caption{Scatterplot of first two PCA dimensions of mean Biber features.}
    \label{fig:source_category_pca_ellipses}
\end{figure*}

In this analysis, we look at the writing style across genres and sources (human or a specific LLM) using the Biber features. For each genre–source pair, we compute the average of each feature, yielding a 67-dimensional feature vector (one dimension per Biber feature). Each feature value represents the weighted average over all text chunks within that genre and source. This results in a matrix $\mathbf{X} \in \mathbb{R}^{96 \times 67}$, corresponding to $8$ genres and $12$ sources (i.e., $8 \times 12 = 96$). We perform z-score normalization on this matrix for each feature to have zero mean and unit standard deviation. We then perform Principal Component Analysis (PCA) and visualize the first two principal components in Figure~\ref{fig:source_category_pca_ellipses}. We further plot confidence ellipses for each category, centered at the mean of the genre and axis lengths are proportional (twice in our implementation) to the standard deviation along those axes. This leads to an interesting observation -- the genre plays a more prominent role in shaping the writing style than the source, i.e., texts from same genre tend to cluster together irrespective of the source. Further, the confidence ellipses are elongated for \texttt{Reviews} and \texttt{Poetry}, indicating a larger variance along the first principal component, which is likely due to creative differences in \texttt{Poetry} and subjectivity in \texttt{Reviews}. Finally, while \texttt{Poetry}, \texttt{Abstracts} cluster away from the center showing more specialized or niche writing styles, genres like \texttt{Books}, \texttt{News} and \texttt{Wikipedia} cluster closer to the center, indicating a more general writing style. 

\begin{figure}[h!]
    \vspace{-2mm}
    \begin{center}
    \includegraphics[width=0.48\textwidth]{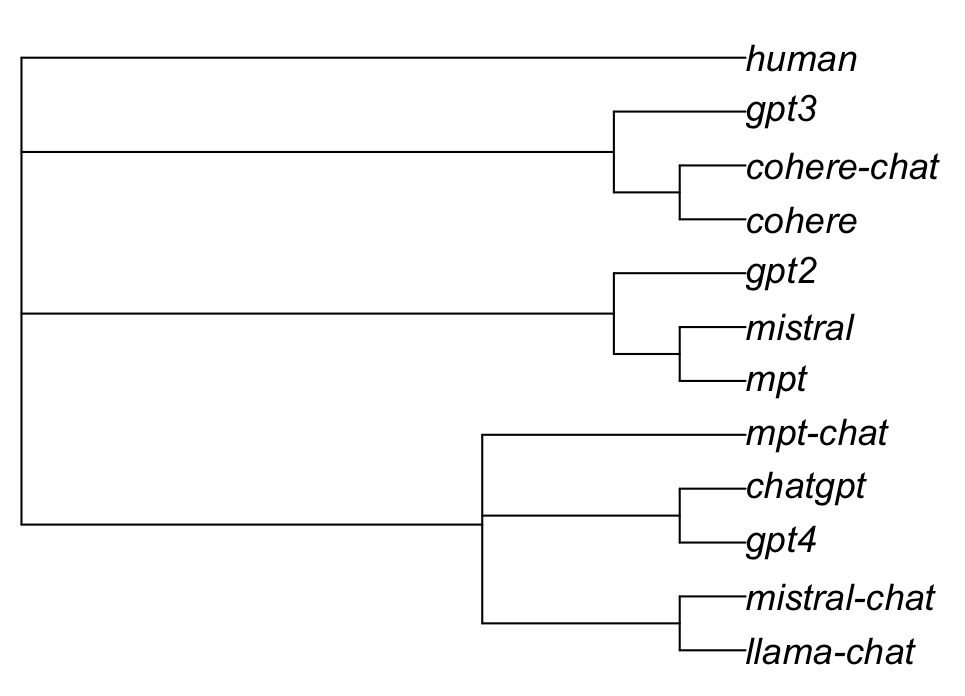}
    \caption{Relaxed consensus tree summarizing hierarchical clustering results over all genres.}
    \label{fig:consensus_cluster}
    \end{center}
    \vspace{-5mm}
\end{figure}

\begin{figure*}[ht]
\vspace{-2mm}
\begin{center}
\includegraphics[width=0.48\textwidth]{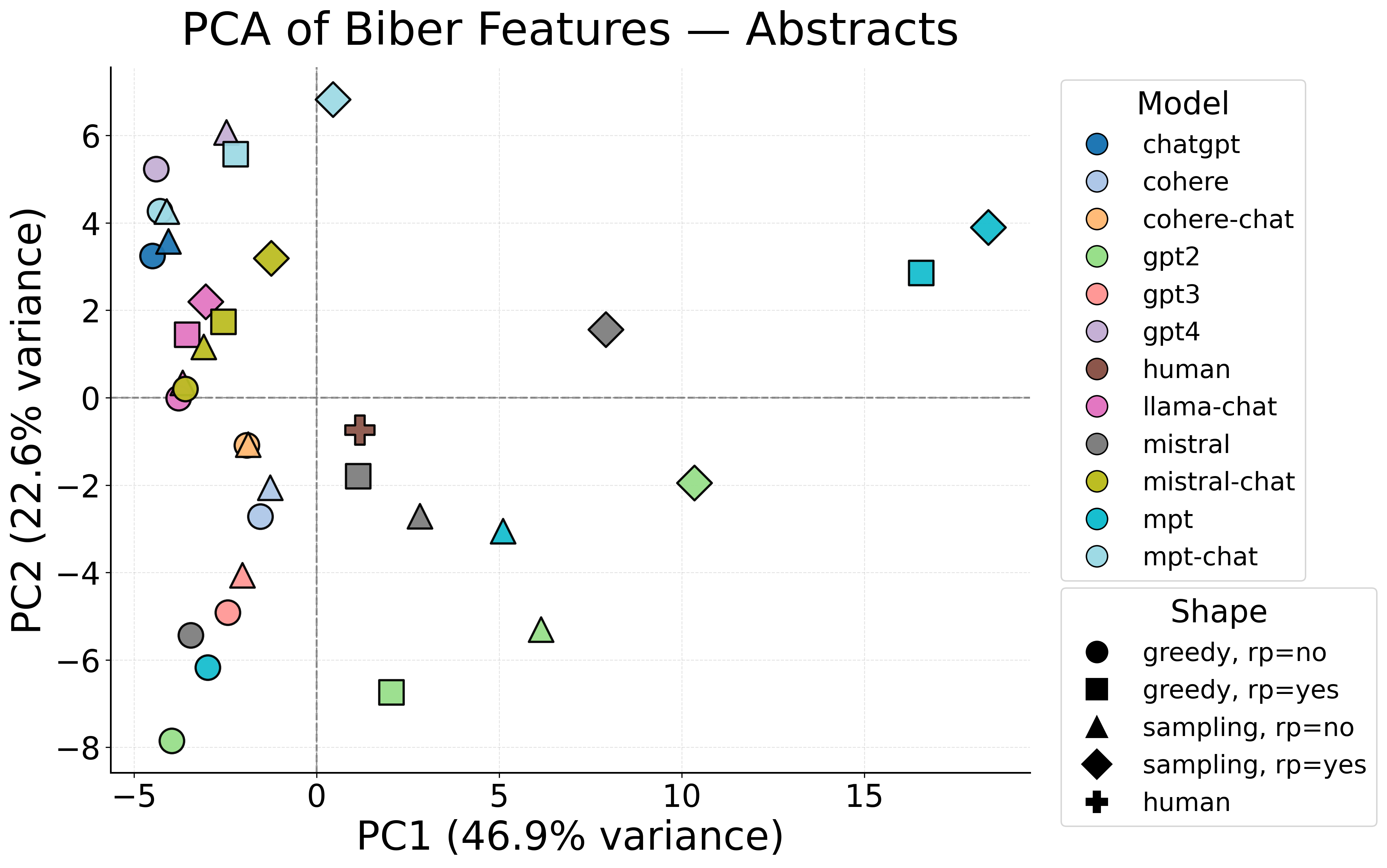}
\hfill
\includegraphics[width=0.48\textwidth]{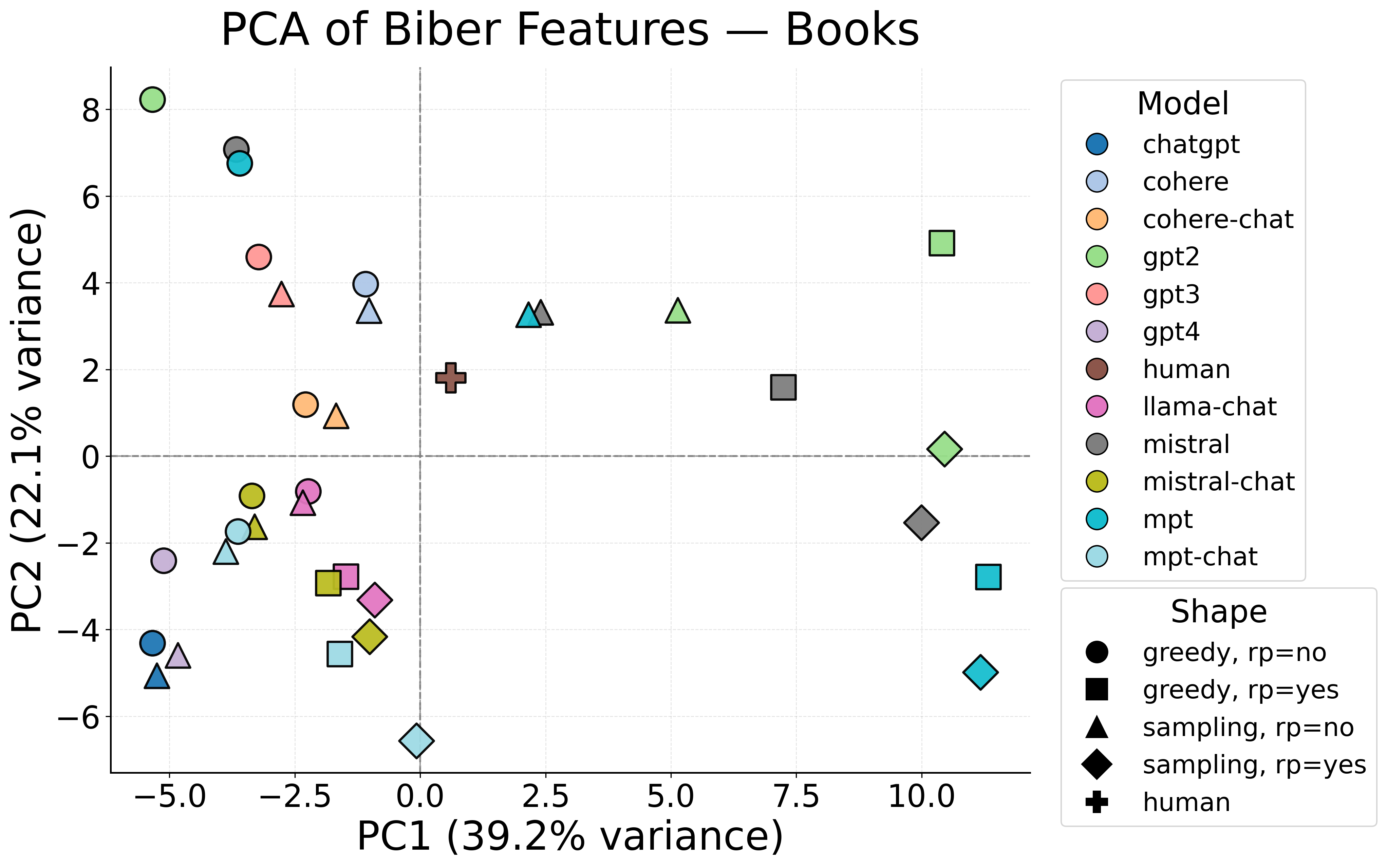}
\vspace{2mm}
\caption{Visualizations of Biber features across selected RAID genres. Colors indicate model identity, while marker shapes indicate decoding and repetition-penalty settings. Human text is shown with a distinct marker.}
\label{fig:decoding_selected}
\end{center}
\vspace{-2mm}
\end{figure*}

\subsection{Hierarchical Clustering of Sources of Text}
\label{ssec:clustering}
We study the similarities between text sources by clustering the 11 LLM-generated text documents together with human-written text. As genre is a dominant factor, we perform clustering separately for each genre to avoid mixing cross-genre and cross-source patterns. For each genre–source pair, we construct a feature vector using the weighted average of Biber features across all text chunks. We then z-score normalize each feature by mean-centering across sources. To reduce redundancy among correlated features, we apply PCA and retain components that explain 95\% of the variance, typically reducing the dimensionality from 67 to 6–7 components. We perform hierarchical clustering on these PCA-reduced representations using Ward’s method\footnote{\url{https://docs.scipy.org/doc/scipy/reference/cluster.hierarchy.html}} and visualize the resulting dendrograms in Appendix Figure~\ref{fig:dendrograms_per_category}. 
To summarize clustering patterns across genres, we construct a consensus tree (Figure~\ref{fig:consensus_cluster})\footnote{\url{https://rdrr.io/cran/ape/man/consensus.html}} using a relaxed threshold $(p = 0.4)$, retaining groups that occur in at least 40\% of the genre-specific trees. This relaxation is to allow more compact and interpretable summary of recurring clustering patterns.
The full set of dendrograms (Appendix Figure~\ref{fig:dendrograms_per_category}) shows that human-written text is frequently grouped with base models. This aligns with findings in~\cite{llm_human_biber_yurko}, suggesting that instruction tuning shifts chat-based models further away from human-written text. We further observe that \textit{Mistral}, \textit{GPT2}, and \textit{MPT} tend to cluster together. Also, \textit{GPT4} and \textit{ChatGPT} form a group, and \textit{Cohere} and \textit{Cohere-Chat} cluster closely. These patterns show the similarity between base models and their chat variants in some of the model families.

\subsection{Writing Styles per LLM Decoding Strategy}
\label{ssec:decoding}
RAID dataset contains outputs from multiple LLMs under different decoding strategies. Specifically,  greedy decoding and random sampling with and without repetition penalties (where applicable). Decoding strategy determines how tokens are selected from a language model's probability distribution. Greedy decoding picks the most likely token at every time step. Random sampling draws from the language model output distribution. When repetition penalty is applied, the probability of previously generated tokens is reduced. In this section, we analyze how Biber feature usage varies across decoding strategies relative to human-written text. As prior analyses indicate that genre is a dominant factor, we conduct this analysis separately for each of the eight genres to isolate the effect of decoding. For each genre, we visualize the first two principal components of z-score normalized Biber features. For brevity we present two genres (one more specialized- \texttt{Abstracts} and one more generic \texttt{Books}) in Figure~\ref{fig:decoding_selected}. All the genres are presented in the Appendix for reference (Figure~\ref{fig:decoding_pca_all_categories}). We make the following observations. First, across all genres, human-written text tends to fall in the center. Second, model generally has a stronger effect than decoding strategy: points corresponding to the same model (color) tend to cluster together, with some exceptions (e.g., points from \textit{GPT-2}, \textit{MPT}, \textit{Mistral} appear dispersed). For these models decoding strategy can have a noticeable effect in some cases. For example, sampling without repetition penalty tends to produce similar patterns, while greedy decoding without repetition penalty also groups together across these models. This may be because these models tend to cluster together irrespective of decoding strategy (Section~\ref{ssec:clustering}).

\section{Conclusion}
\label{sec:conclude}
In this work, we analyze the RAID dataset, a large-scale benchmark for machine-generated text detection, using Biber features to characterize the stylistic variation between machine-generated and human-written text across multiple models, genres and decoding strategies. Our findings can be summarized as follows. First, the dominant stylistic patterns in LLM-generated text appear robust to variations in prompt settings. Second, genre plays a stronger role in shaping stylistic features than the source of text itself, showing that LLMs have effectively learned to mimic the genre-specific variation in human writing. Third, the chat versions of the models tend to cluster together indicating that instruction tuning and reinforcement learning push models toward similar stylistic tendencies. Finally, within each genre, model identity generally has a stronger influence on stylistic features than decoding strategy, with some exceptions. Understanding these stylistic differences is important for several reasons. It can inform more effective use of LLMs, for example by guiding prompting strategies to achieve desired writing styles and also on the other hand, identify characteristic signatures of machine-generated text, aiding detection. As LLMs continue to evolve, ongoing analysis will be necessary to track how their stylistic characteristics change over time.

\section*{Copyright and Acknowledgments}
Copyright 2026 Carnegie Mellon University.

This material is based upon work supported by the Department of War under Air Force Contract No. FA8702-15-D-0002 with Carnegie Mellon University for the operation of the Software Engineering Institute, a federally funded research and development center.  

The opinions, findings, conclusions, and/or recommendations contained in this material are those of the author(s) and should not be construed as an official US Government position, policy, or decision, unless designated by other documentation.  

References herein to any specific entity, product, process, or service by trade name, trademark, manufacturer, or otherwise, does not necessarily constitute or imply its endorsement, recommendation, or favoring by Carnegie Mellon University or its Software Engineering Institute nor of Carnegie Mellon University - Software Engineering Institute by any such named or represented entity.

NO WARRANTY. THIS CARNEGIE MELLON UNIVERSITY AND SOFTWARE ENGINEERING INSTITUTE MATERIAL IS FURNISHED ON AN "AS-IS" BASIS. CARNEGIE MELLON UNIVERSITY MAKES NO WARRANTIES OF ANY KIND, EITHER EXPRESSED OR IMPLIED, AS TO ANY MATTER INCLUDING, BUT NOT LIMITED TO, WARRANTY OF FITNESS FOR PURPOSE OR MERCHANTABILITY, EXCLUSIVITY, OR RESULTS OBTAINED FROM USE OF THE MATERIAL. CARNEGIE MELLON UNIVERSITY DOES NOT MAKE ANY WARRANTY OF ANY KIND WITH RESPECT TO FREEDOM FROM PATENT, TRADEMARK, OR COPYRIGHT INFRINGEMENT.

[DISTRIBUTION STATEMENT A] This material has been approved for public release and unlimited distribution.  Please see Copyright notice for non-US Government use and distribution.

This work is licensed under a Creative Commons Attribution-NonCommercial 4.0 International License (https://creativecommons.org/licenses/by-nc/4.0/).  Requests for permission for non-licensed uses should be directed to the Software Engineering Institute at permission@sei.cmu.edu.

This work product was created in part using generative AI.

DM26-0386

\bibliography{metrics-bib}

\appendix
\section{Appendix}
\label{sec:appendix}

\begin{figure}[ht]
    \centering
    \includegraphics[width=0.48\textwidth]{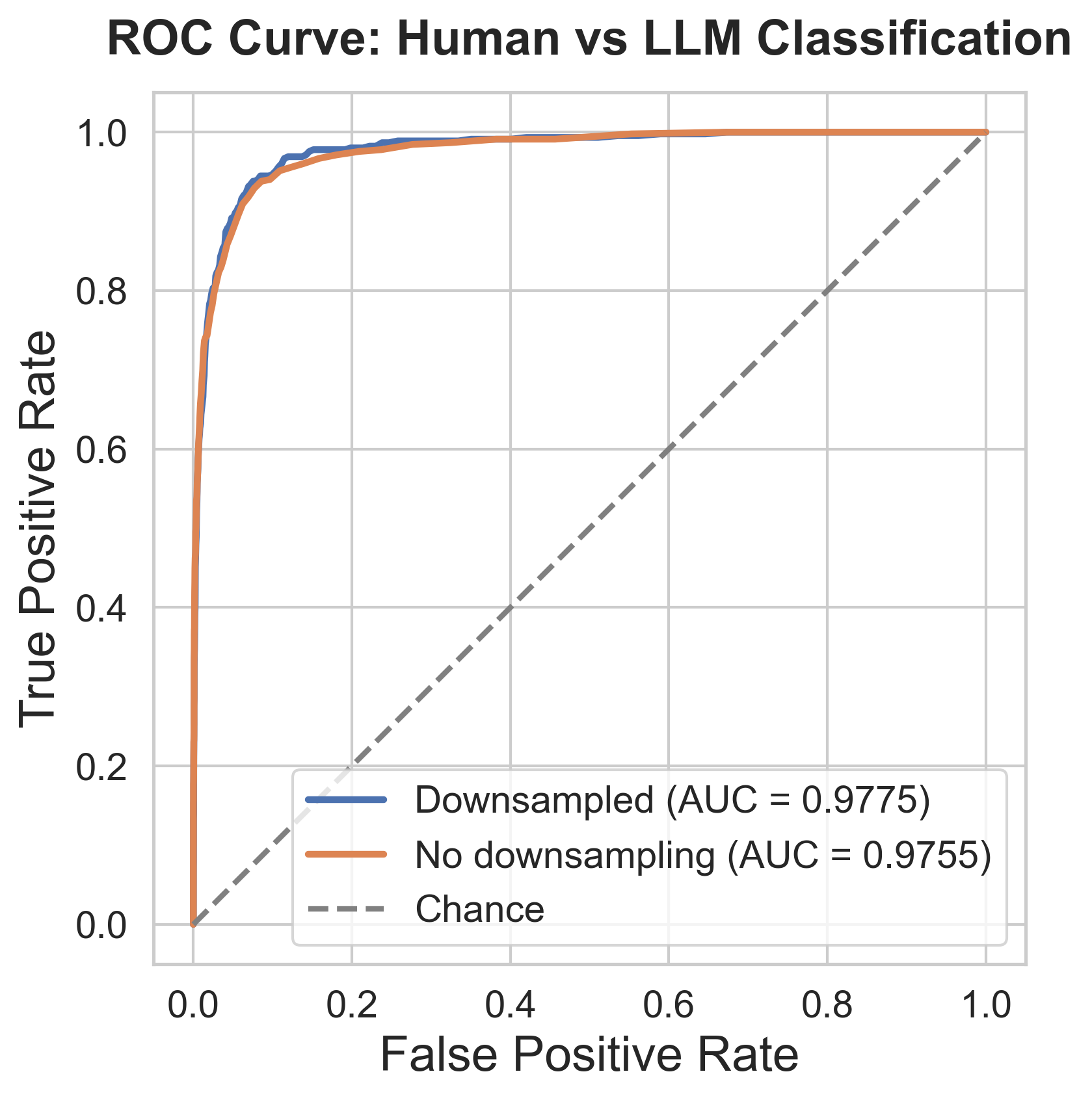}
    \caption{ROC curve comparing base model and downsampled model.}
    \label{fig:roc}
\end{figure}

\begin{figure*}[t]
    \centering
    \includegraphics[width=\textwidth]{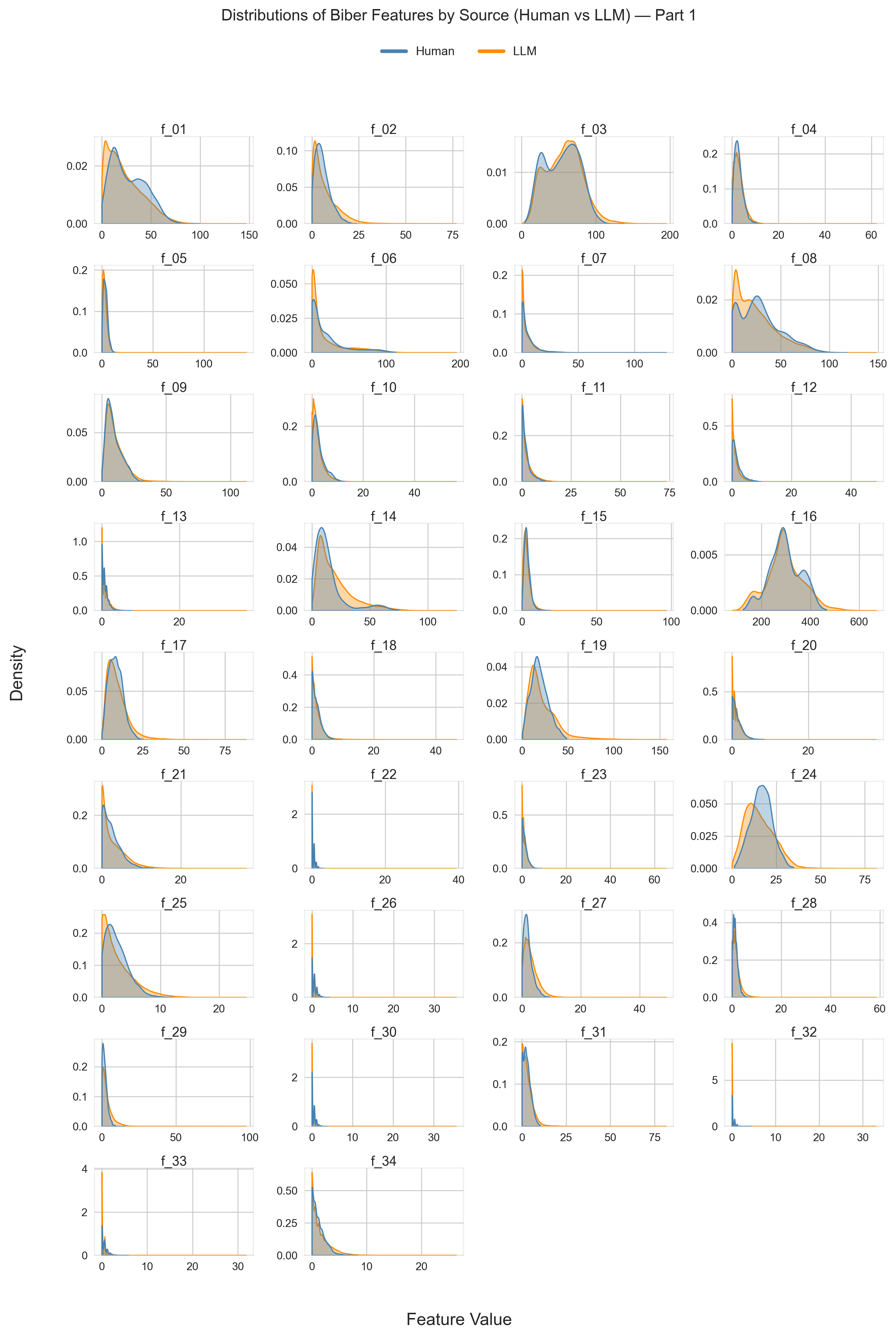}
    \caption{Distribution of Biber features (Part 1).}
    \label{fig:biber_dist_part1}
\end{figure*}

\begin{figure*}[t]
    \centering
    \includegraphics[width=\textwidth]{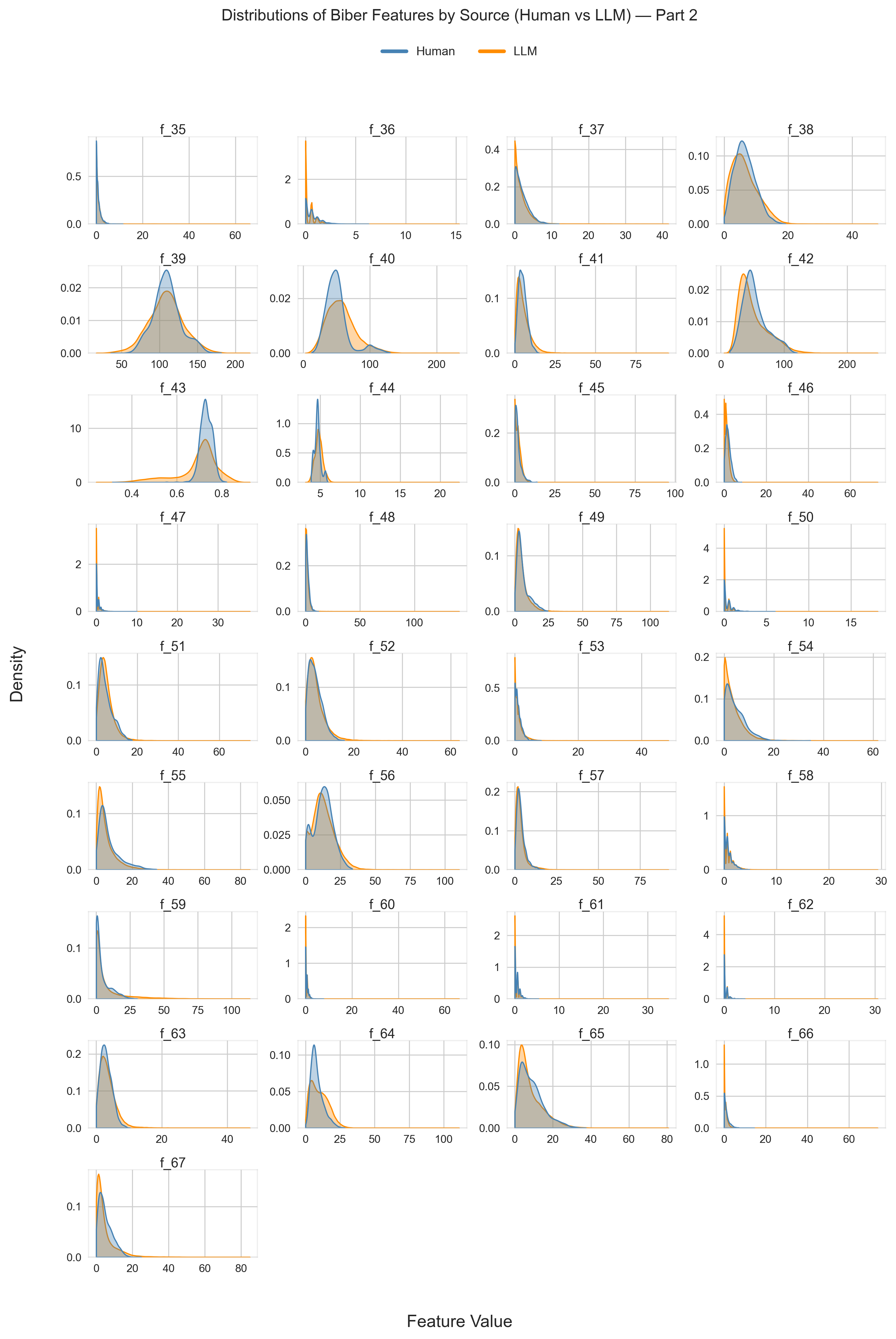}
    \caption{Distribution of Biber features (Part 2).}
    \label{fig:biber_dist_part2}
\end{figure*}

\begin{figure*}[htbp]
    \centering
    \includegraphics[width=0.85\textwidth]{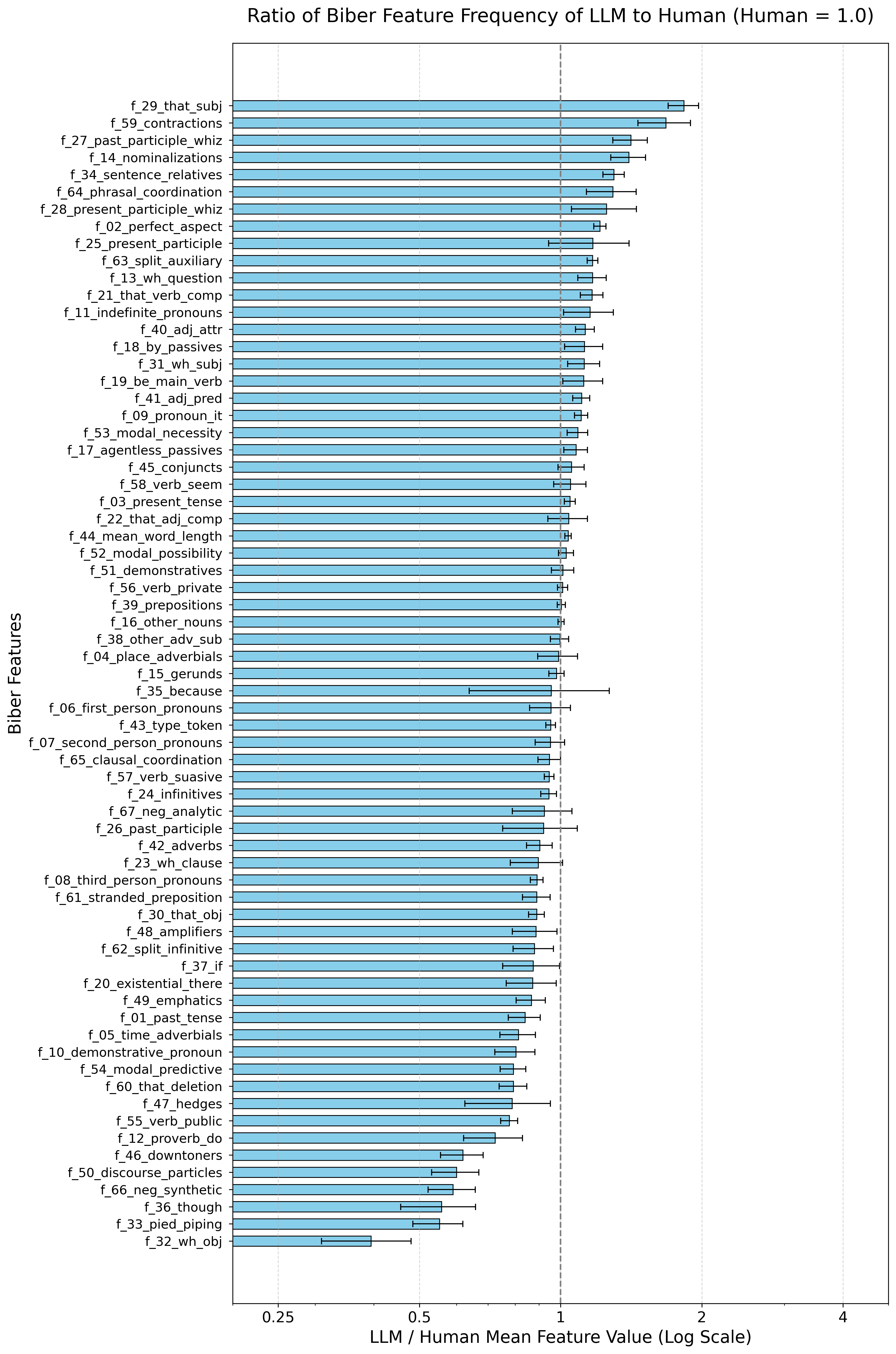}
    \caption{Ratio of Feature Usage LLM Vs. Human for All Features.}
    \label{fig:ratio_plot_all}
\end{figure*}

\begin{figure*}[htbp]
    \centering
    \includegraphics[width=0.85\textwidth]{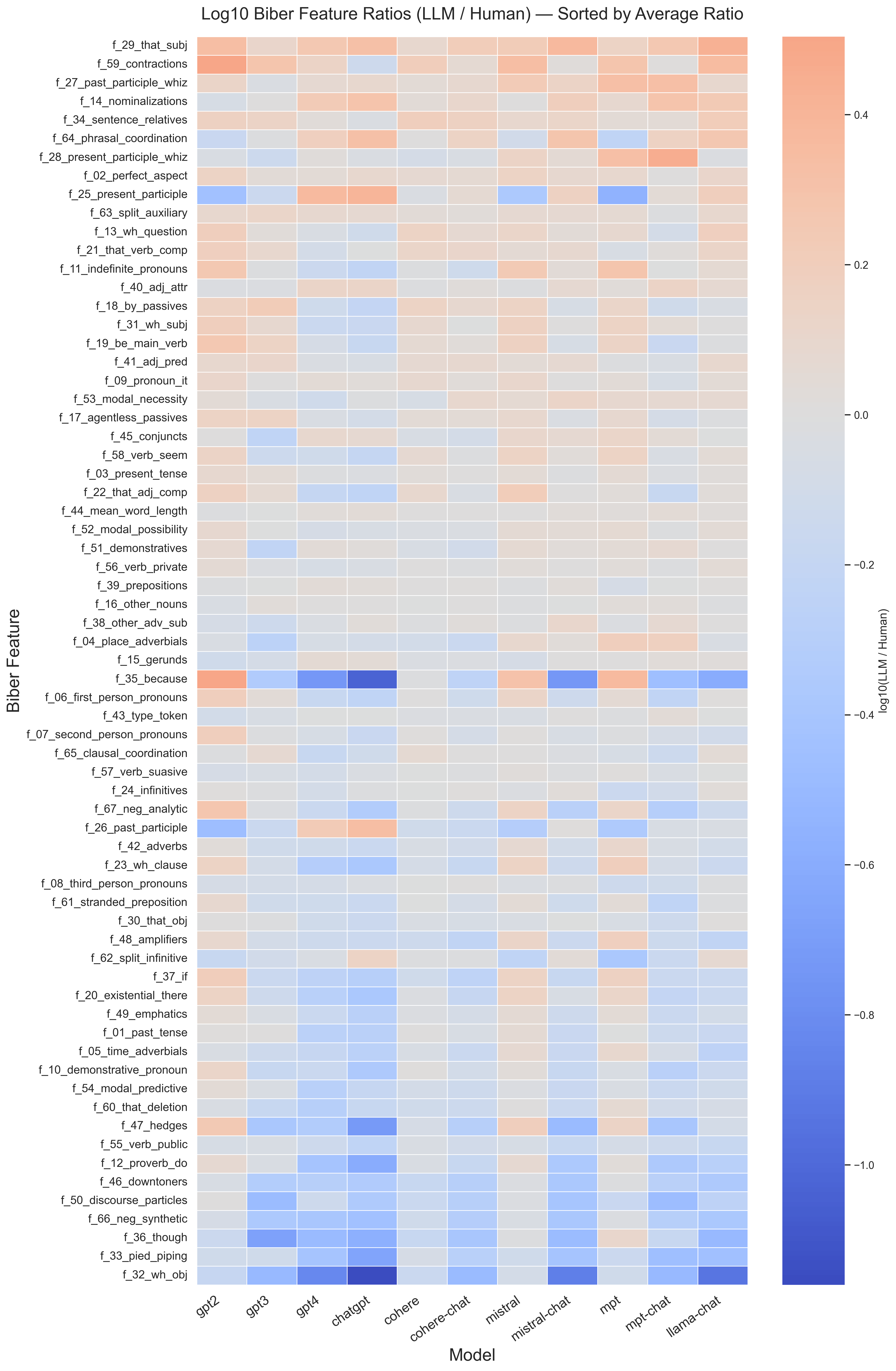}
    \caption{Heat map of Log10 Biber Feature Ratios (LLM / Human) for All Features}
    \label{fig:heat_map_all}
\end{figure*}

\begin{figure*}[htbp]
    \centering

    % Top-left: Raw Features
    \begin{subfigure}[b]{0.48\textwidth}
        \includegraphics[width=\linewidth]{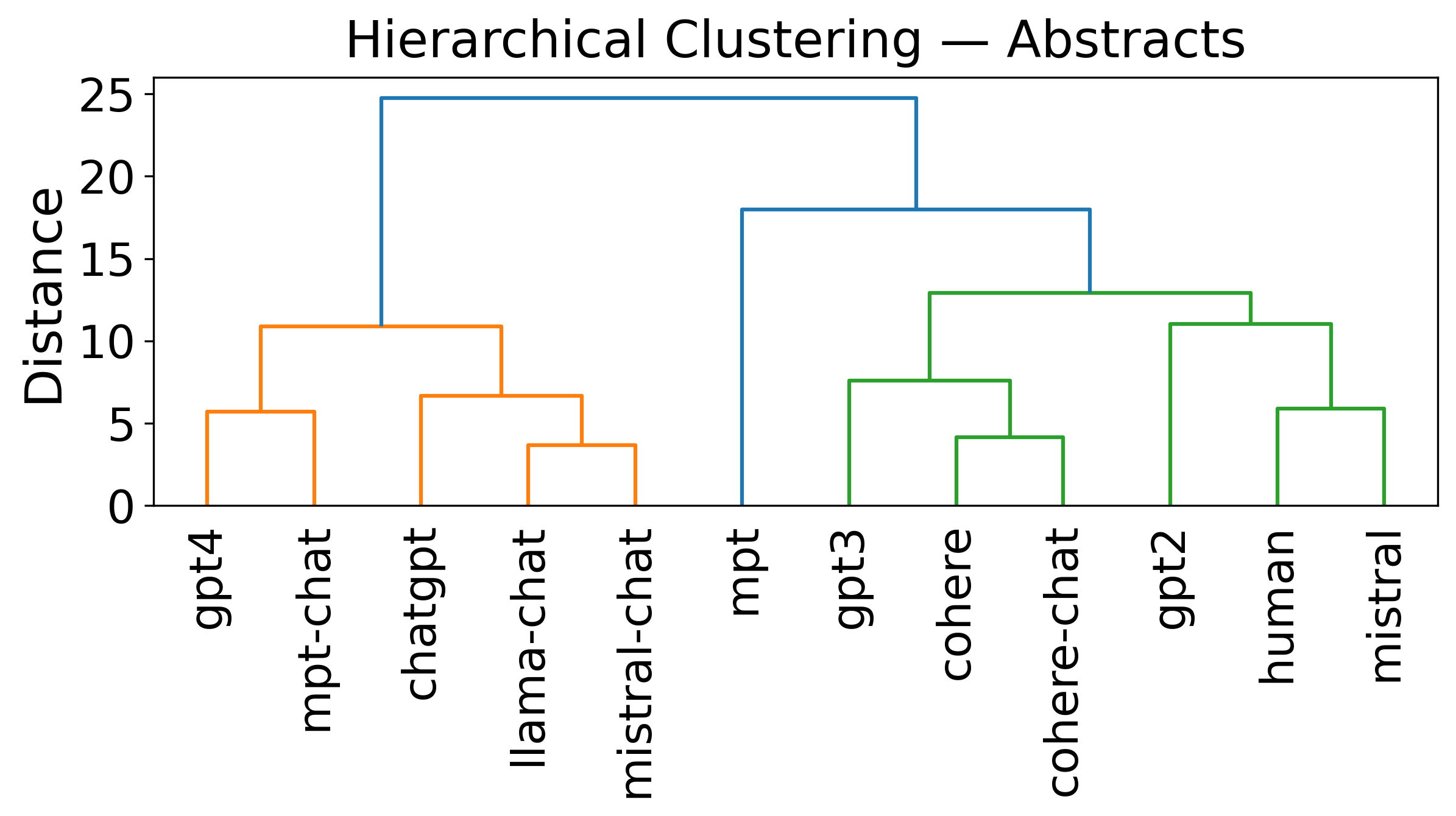}
        \caption{Abstracts}
    \end{subfigure}
    \hfill
    % Top-right: Normalized
    \begin{subfigure}[b]{0.48\textwidth}
        \includegraphics[width=\linewidth]{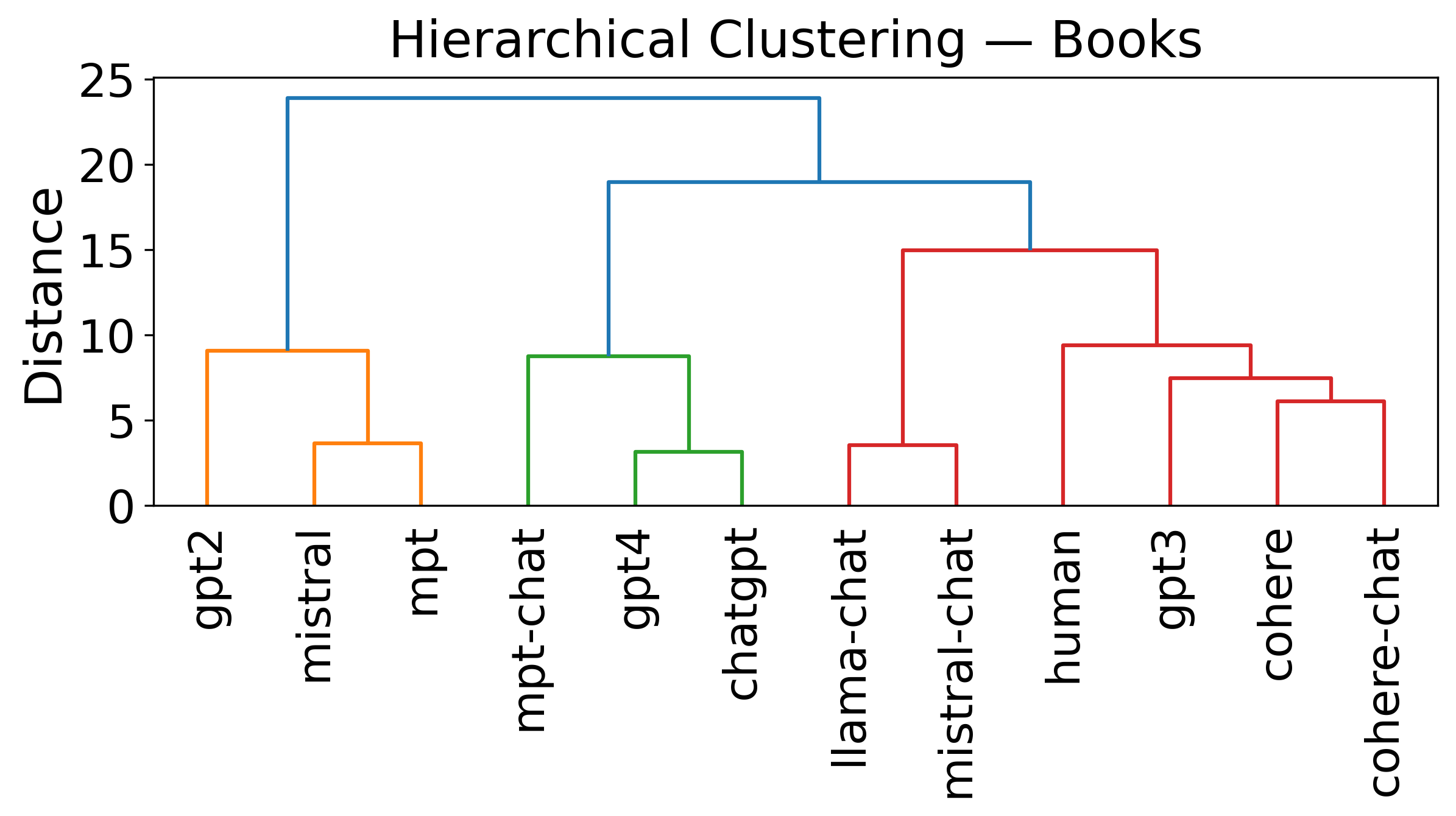}
        \caption{Books}
    \end{subfigure}

    \vspace{0.5cm}

    % Bottom-left: PCA
    \begin{subfigure}[b]{0.48\textwidth}
        \includegraphics[width=\linewidth]{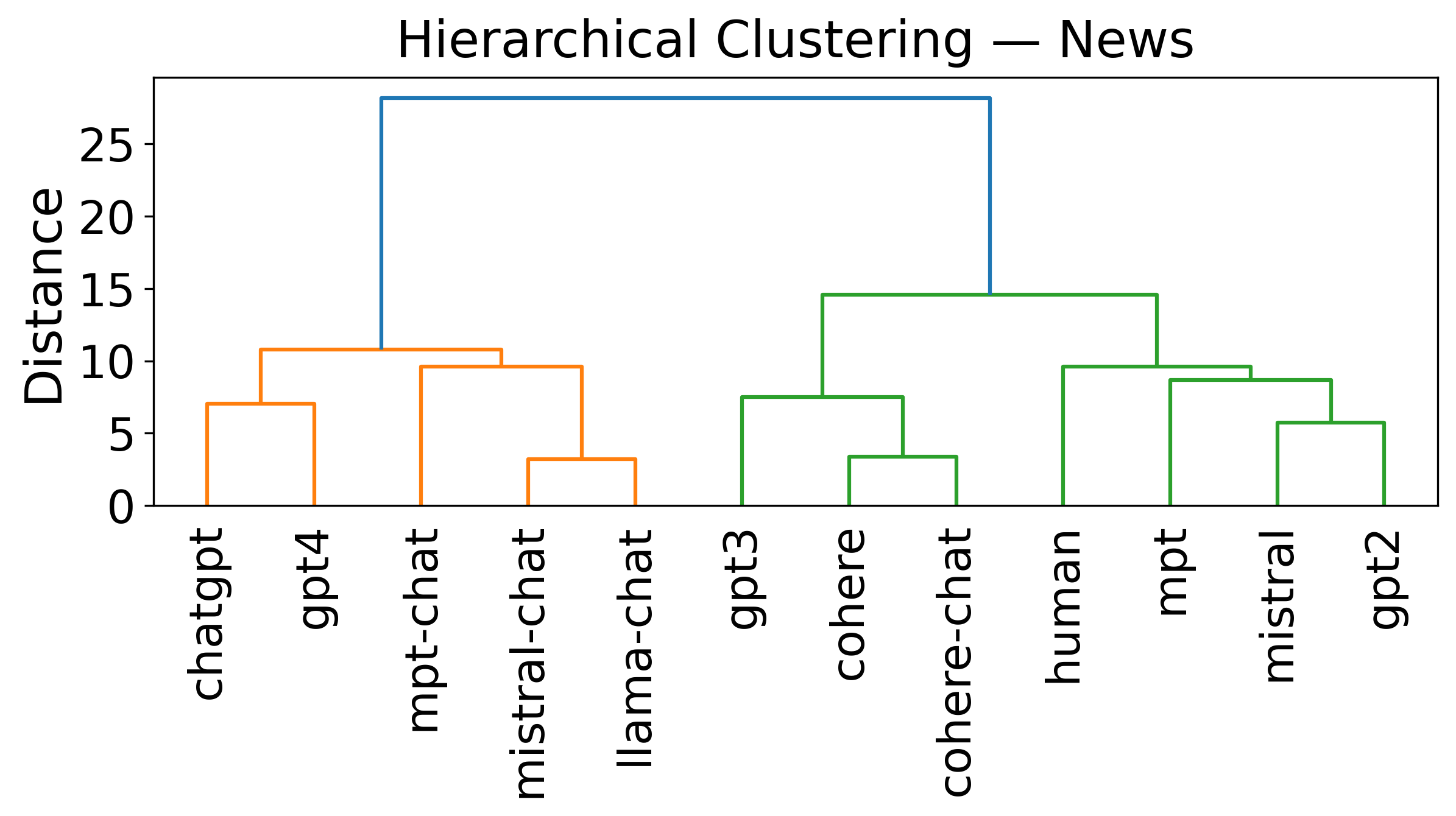}
        \caption{News}
    \end{subfigure}
    \hfill
    % Bottom-right: Important Features
    \begin{subfigure}[b]{0.48\textwidth}
        \includegraphics[width=\linewidth]{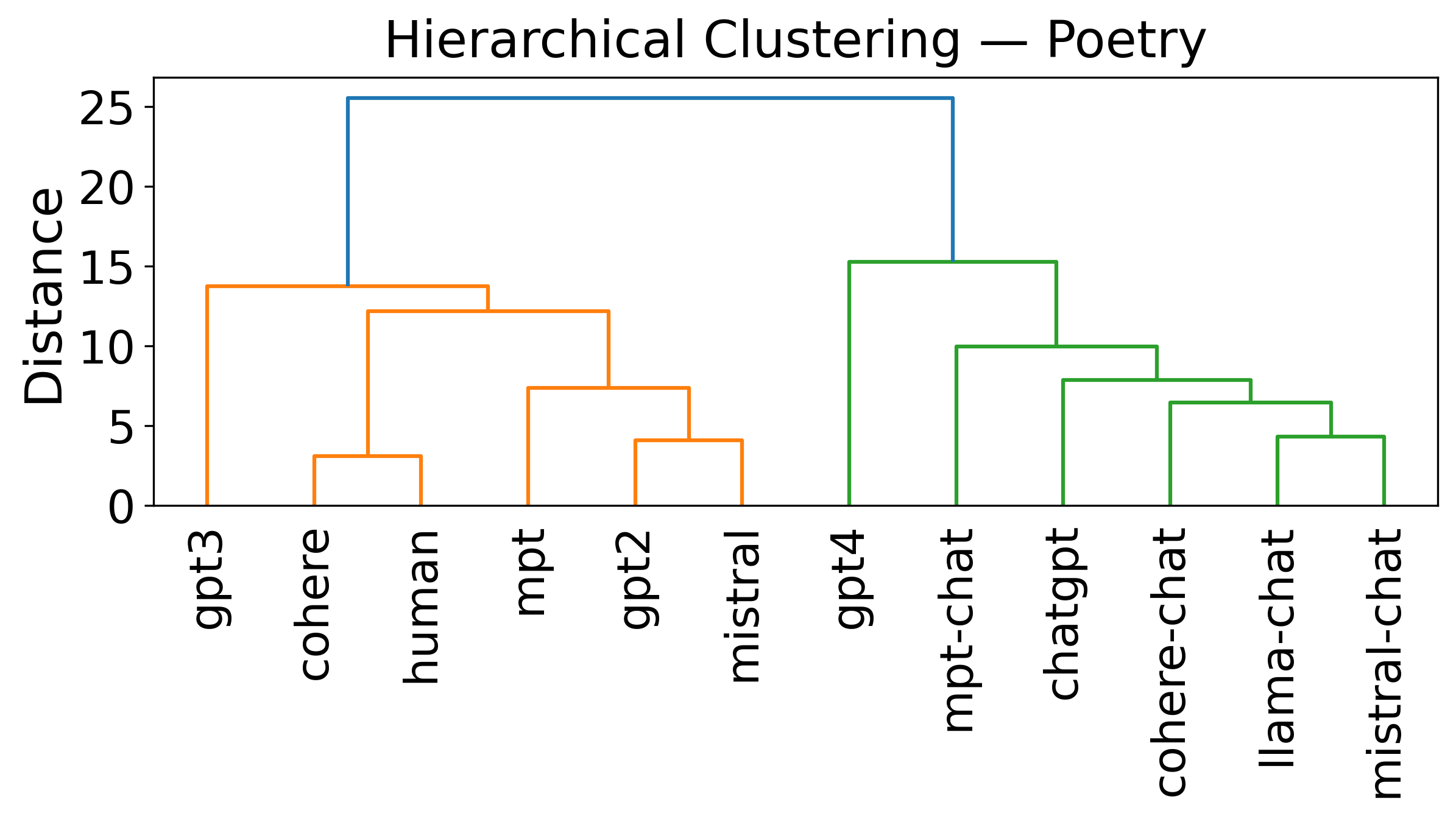}
        \caption{Poetry}
    \end{subfigure}

    \vspace{0.5cm}

    \begin{subfigure}[b]{0.48\textwidth}
        \includegraphics[width=\linewidth]{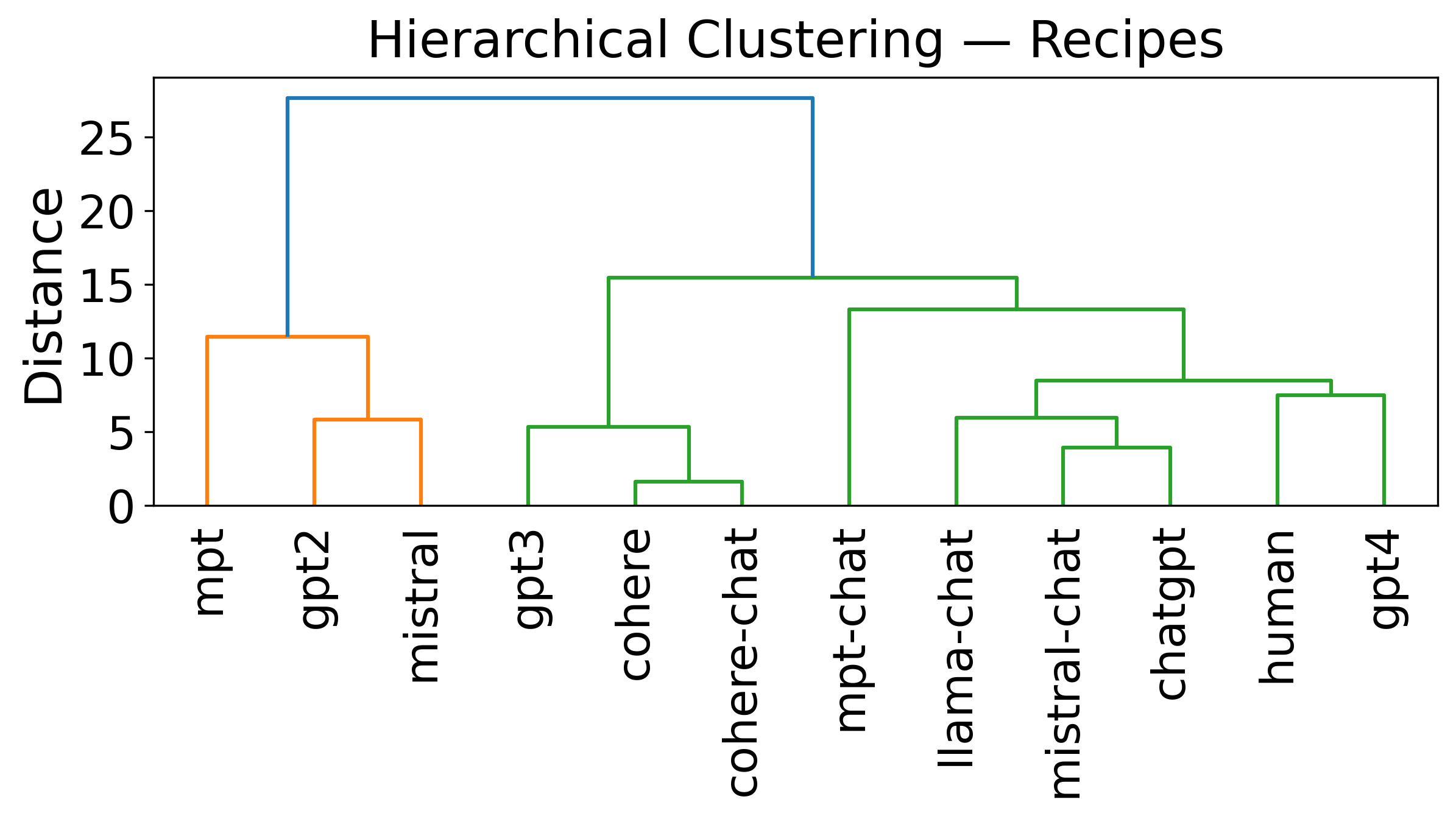}
        \caption{Recipes}
    \end{subfigure}
    \hfill
    \begin{subfigure}[b]{0.48\textwidth}
        \includegraphics[width=\linewidth]{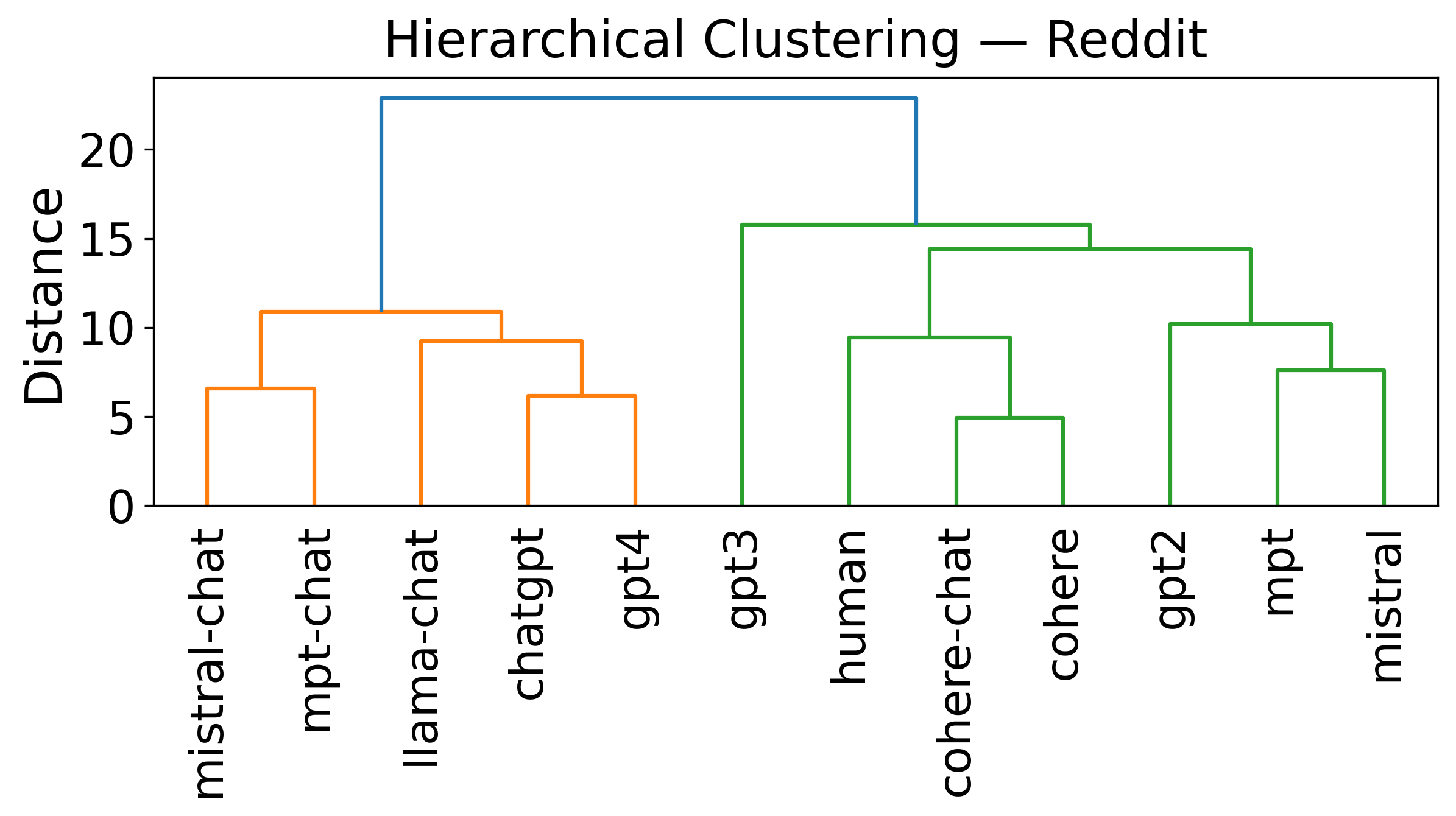}
        \caption{Reddit}
    \end{subfigure}

    \vspace{0.5cm}

    \begin{subfigure}[b]{0.48\textwidth}
        \includegraphics[width=\linewidth]{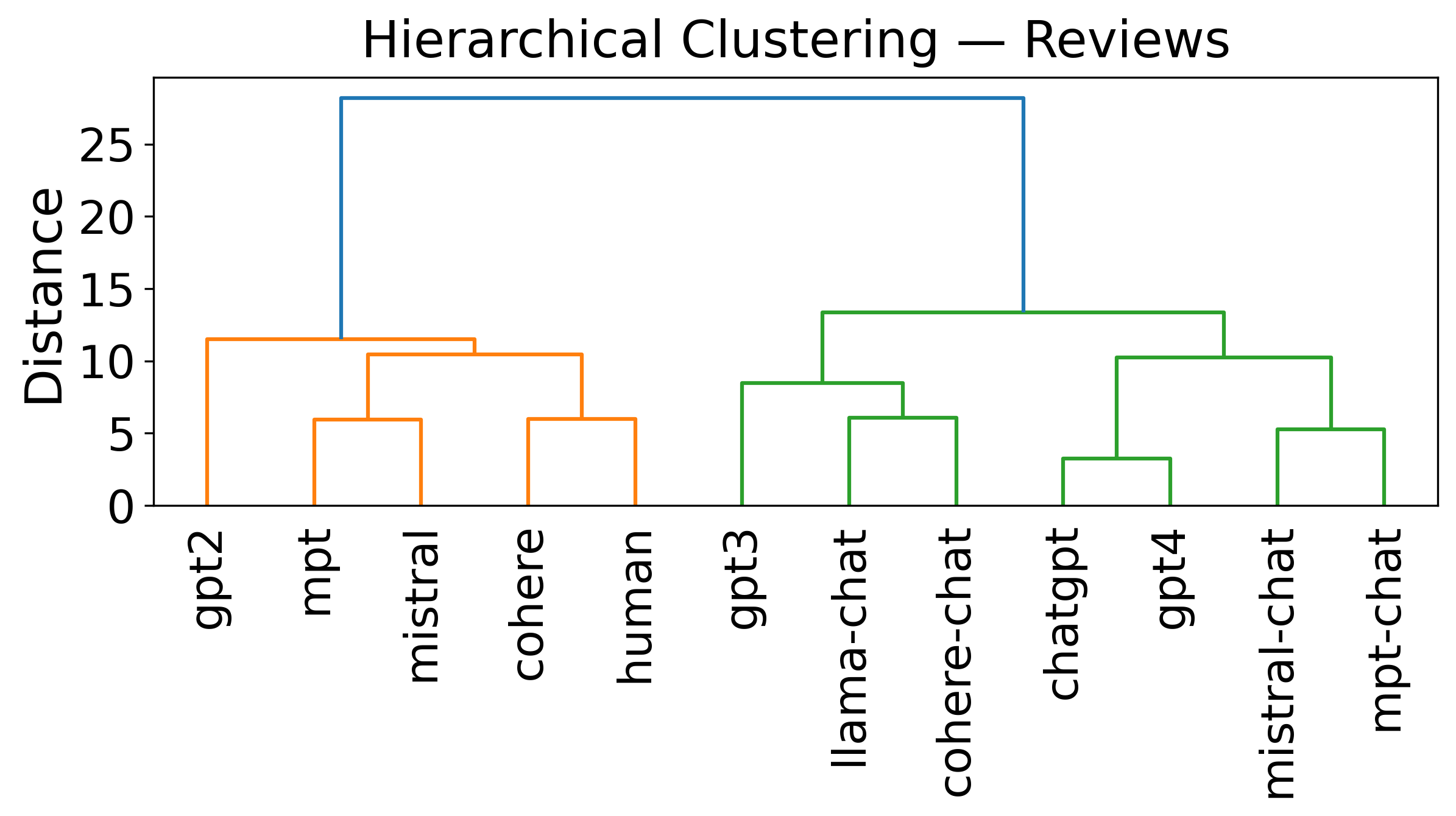}
        \caption{Reviews}
    \end{subfigure}
    \hfill
    \begin{subfigure}[b]{0.48\textwidth}
        \includegraphics[width=\linewidth]{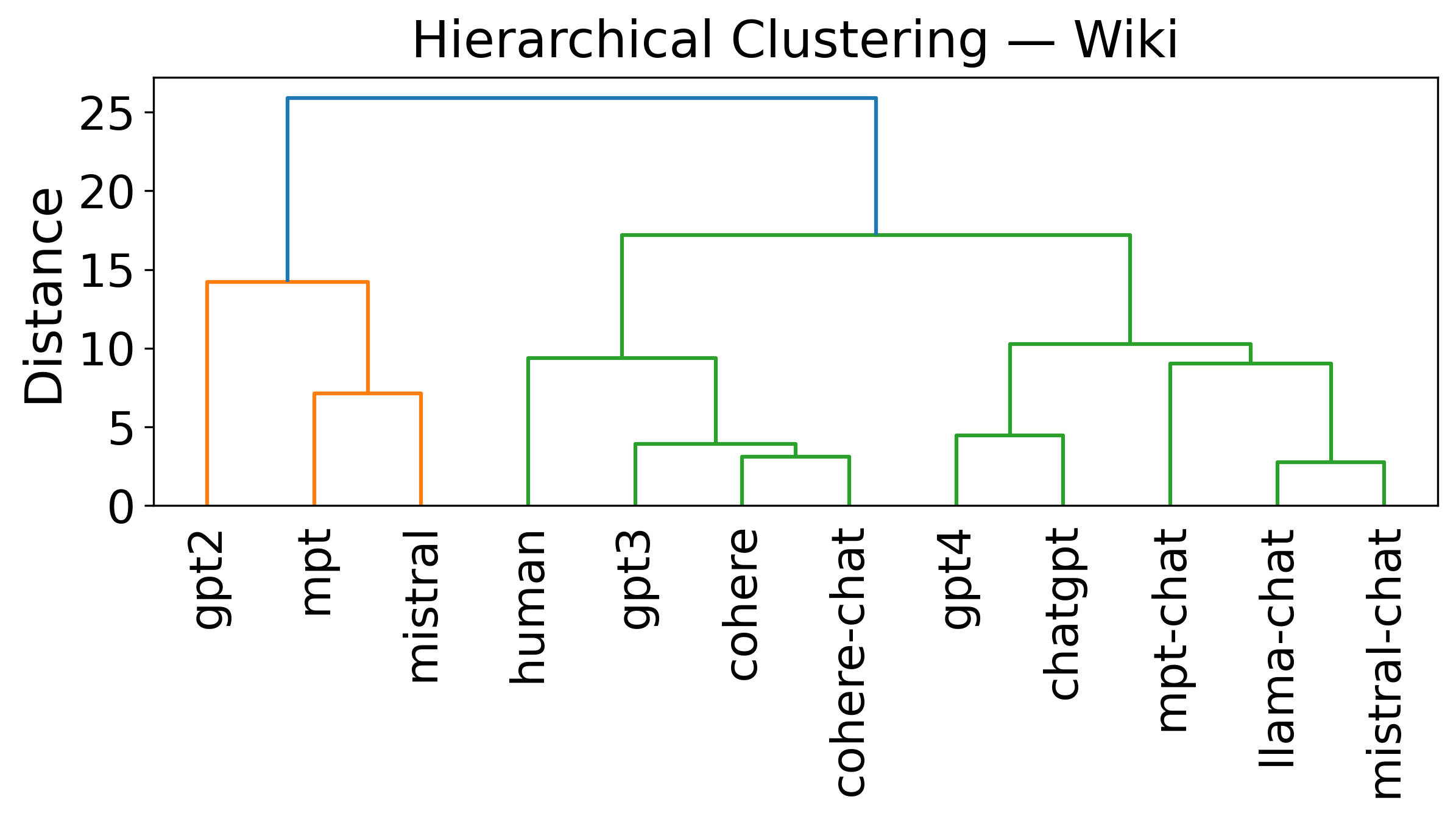}
        \caption{Wikipedia}
    \end{subfigure}

    \caption{Hierarchical clustering dendrograms using PCA with components capturing 95\% variance on normalized biber features.}
    \label{fig:dendrograms_per_category}
\end{figure*}

\begin{figure*}[t]
    \centering

    \begin{subfigure}[t]{0.48\textwidth}
        \centering
        \includegraphics[width=\linewidth]{pca_abstracts_weighted_biber_features_publication.png}
        \caption{Abstracts}
    \end{subfigure}
    \hfill
    \begin{subfigure}[t]{0.48\textwidth}
        \centering
        \includegraphics[width=\linewidth]{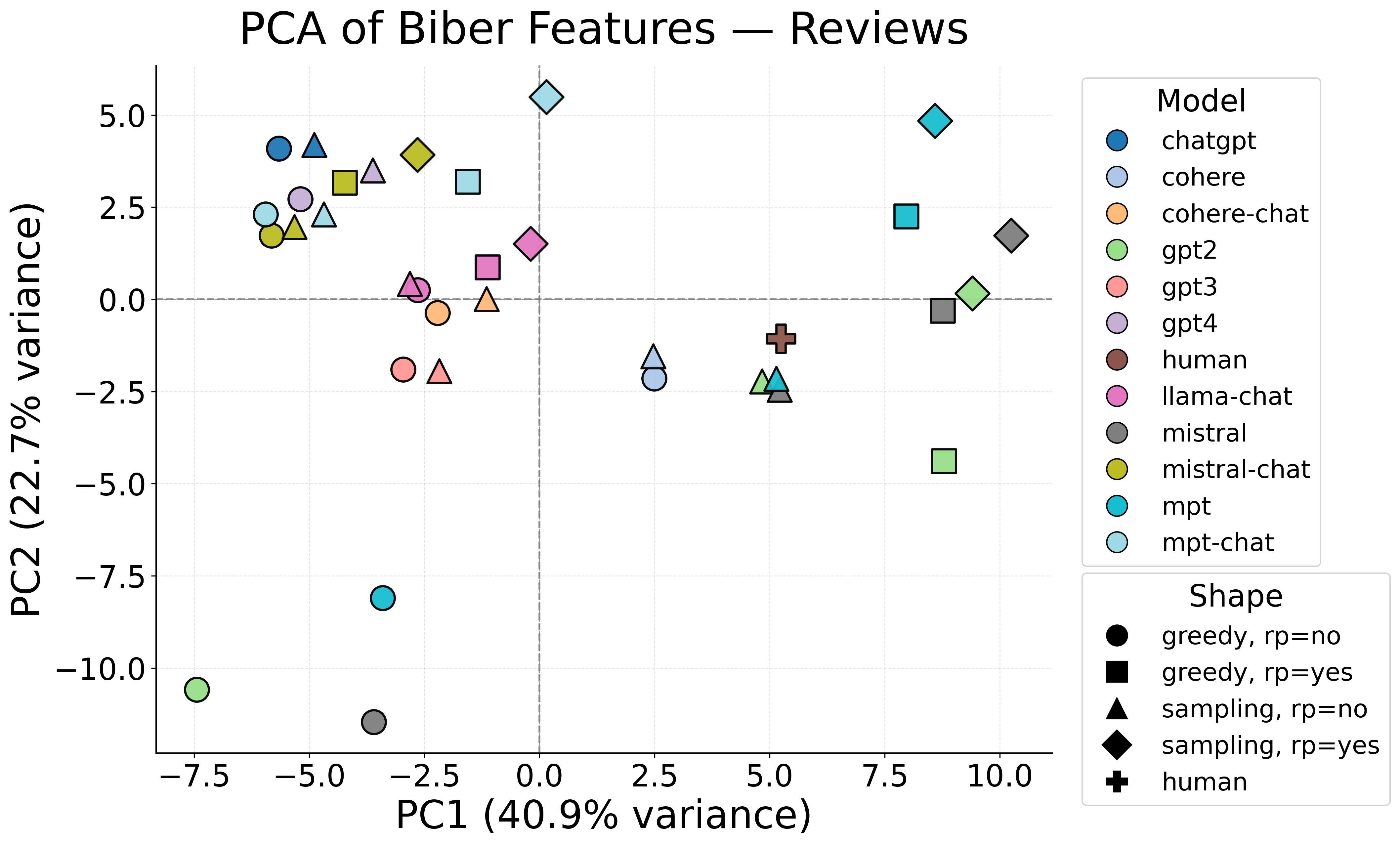}
        \caption{Reviews}
    \end{subfigure}
    \hfill
    \begin{subfigure}[t]{0.48\textwidth}
        \centering
        \includegraphics[width=\linewidth]{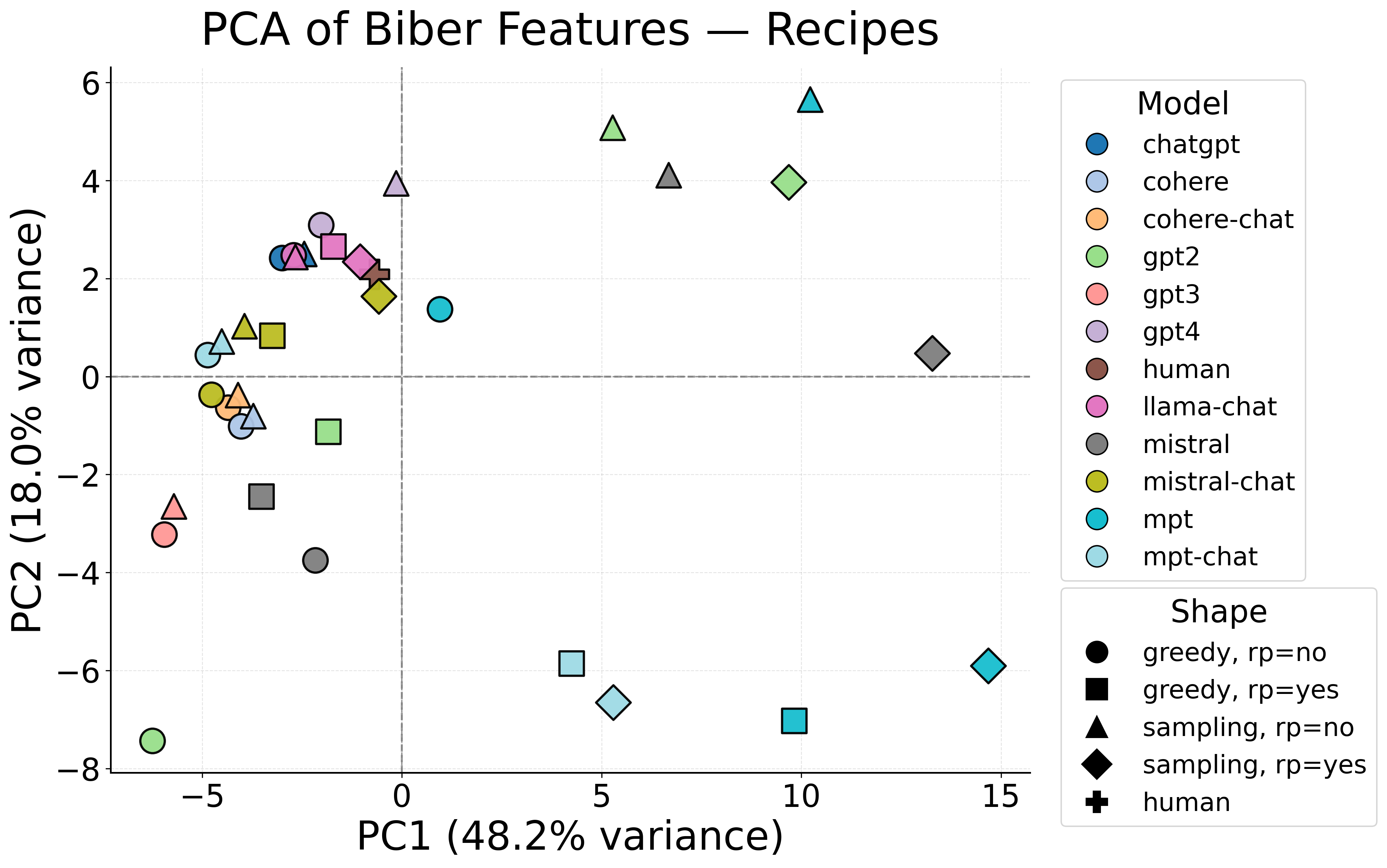}
        \caption{Recipes}
    \end{subfigure}
    \hfill
    \begin{subfigure}[t]{0.48\textwidth}
        \centering
        \includegraphics[width=\linewidth]{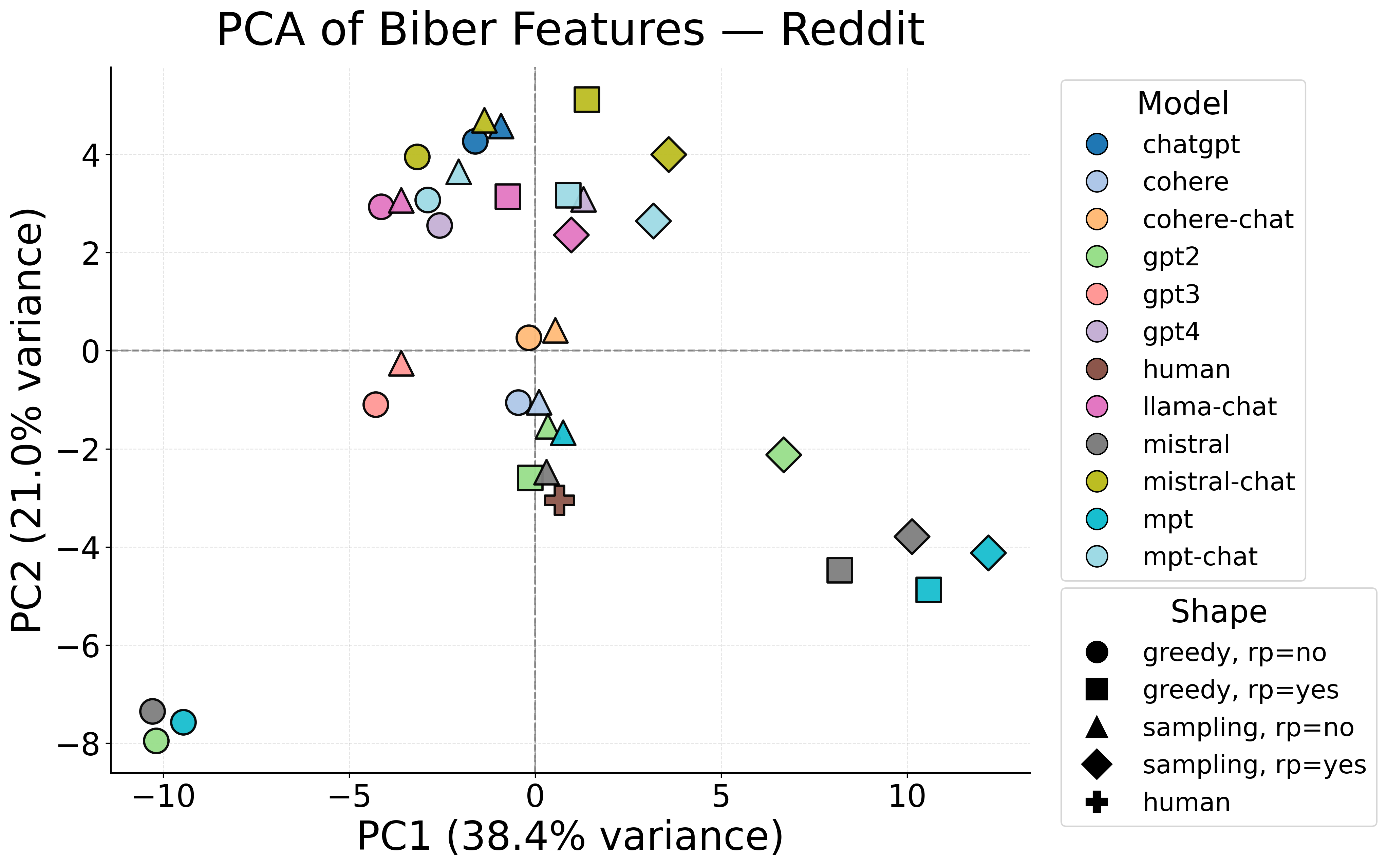}
        \caption{Reddit}
    \end{subfigure}

    \vspace{0.5em}

    \begin{subfigure}[t]{0.48\textwidth}
        \centering
        \includegraphics[width=\linewidth]{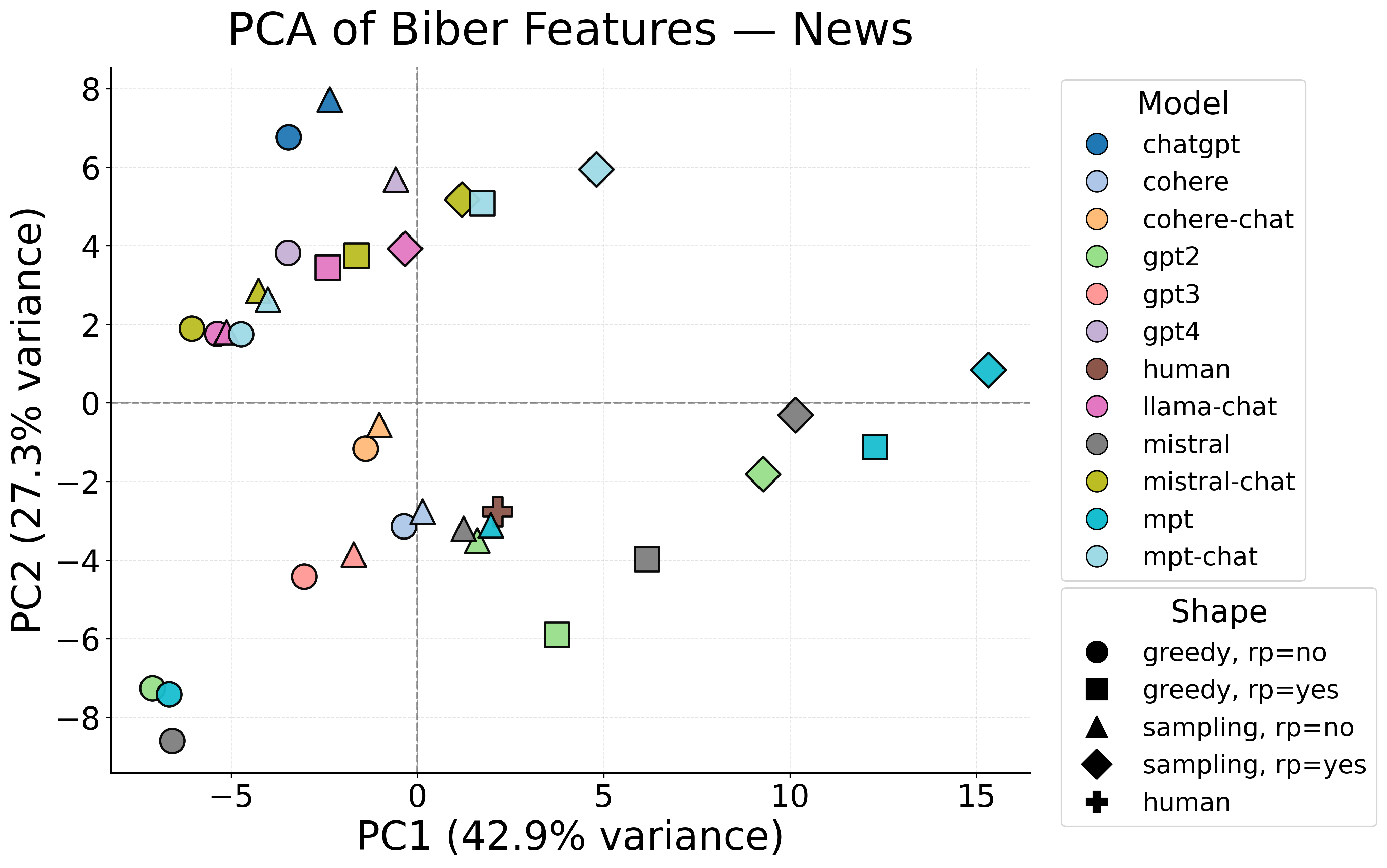}
        \caption{News}
    \end{subfigure}
    \hfill
    \begin{subfigure}[t]{0.48\textwidth}
        \centering
        \includegraphics[width=\linewidth]{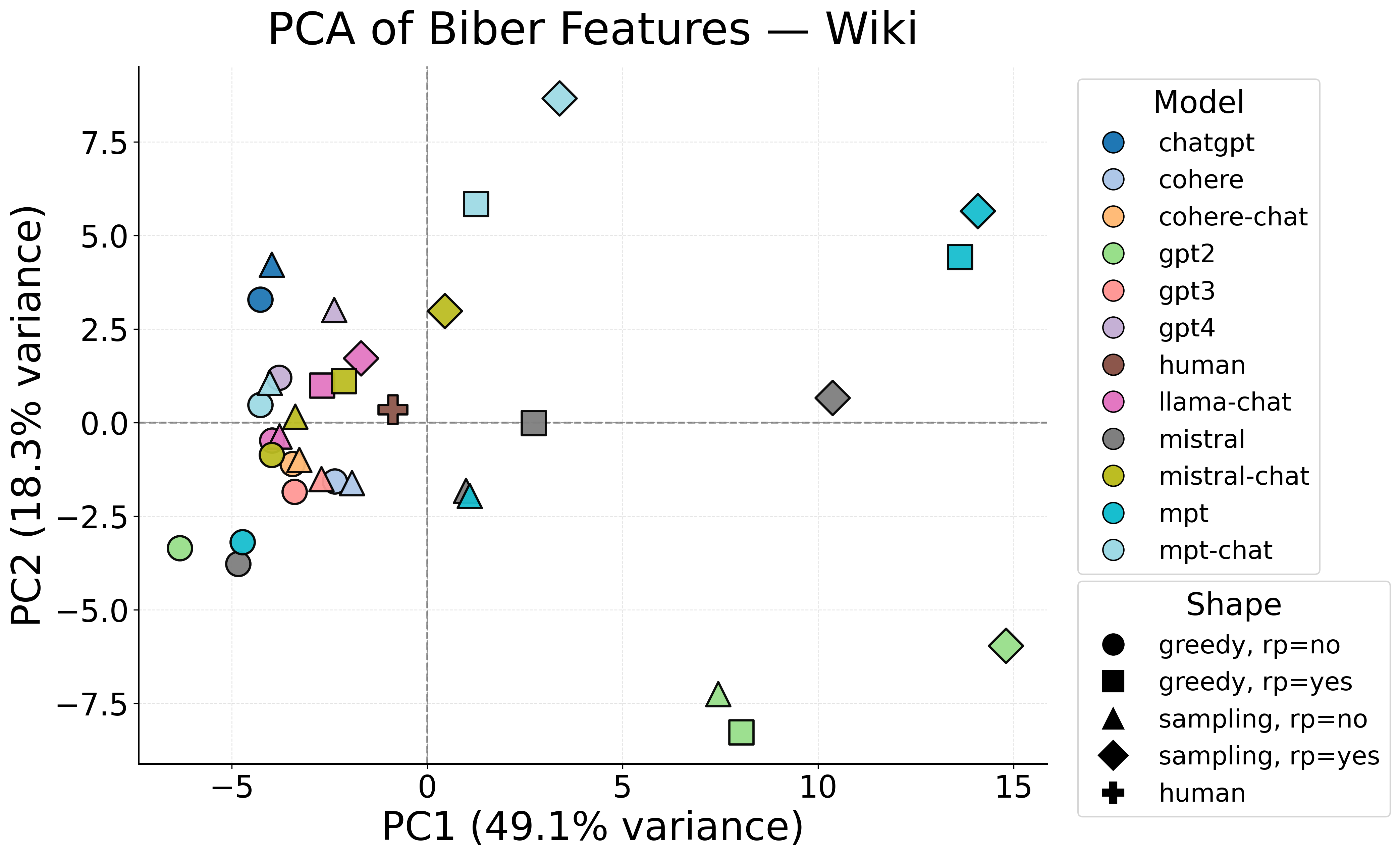}
        \caption{Wiki}
    \end{subfigure}
    \hfill
    \begin{subfigure}[t]{0.48\textwidth}
        \centering
        \includegraphics[width=\linewidth]{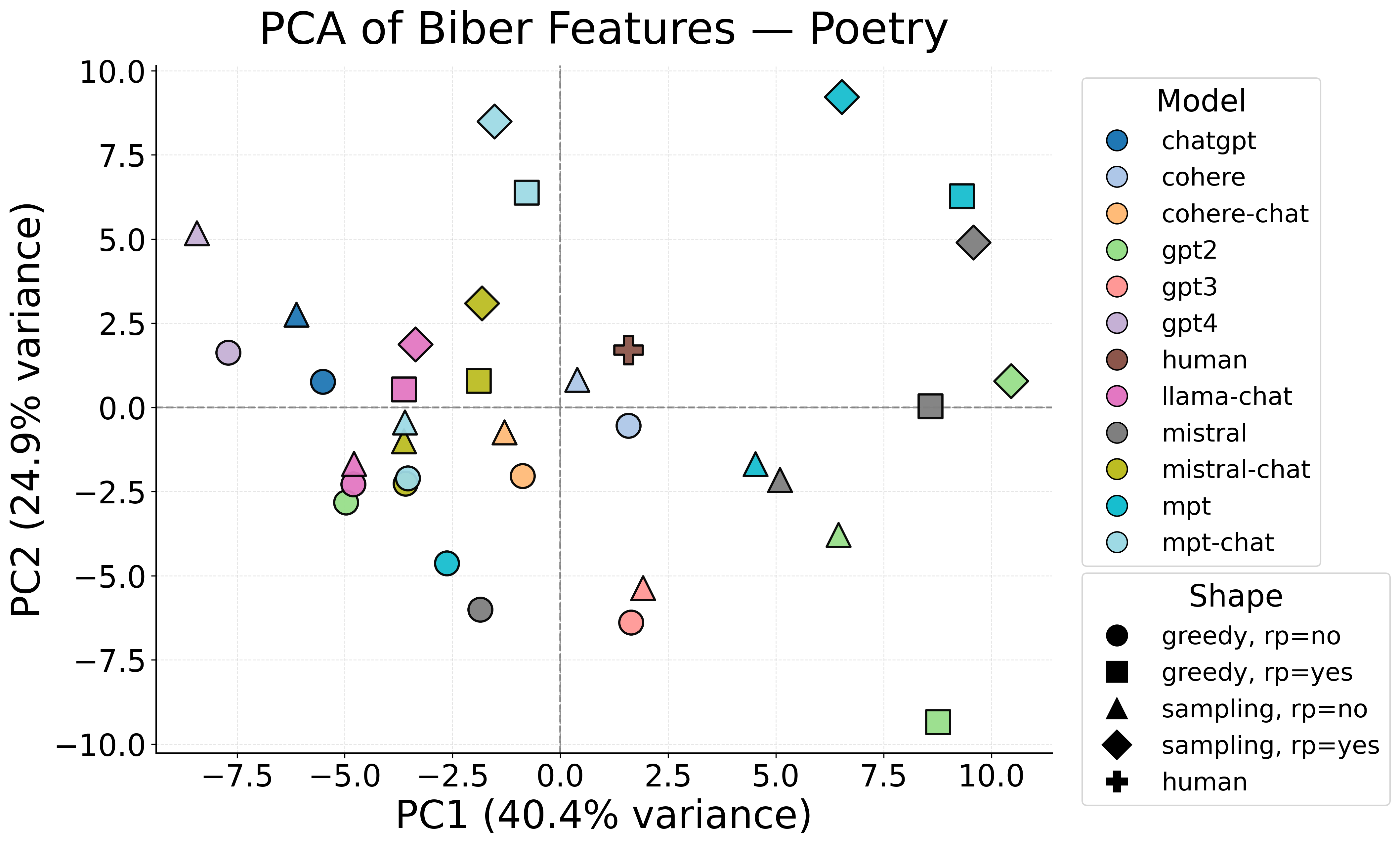}
        \caption{Poetry}
    \end{subfigure}
    \hfill
    \begin{subfigure}[t]{0.48\textwidth}
        \centering
        \includegraphics[width=\linewidth]{pca_books_weighted_biber_features_publication.png}
        \caption{Books}
    \end{subfigure}

    \caption{PCA visualizations of Biber features across eight RAID categories. Colors indicate model identity, while marker shapes indicate decoding and repetition-penalty settings. Human text is shown with a distinct marker.}
    \label{fig:decoding_pca_all_categories}
\end{figure*}

This section provides additional visualizations that support and extend the results presented in the main text. Figure~\ref{fig:roc} shows the ROC curve comparing the base model and the downsampled model explained in Section~\ref{sssec:classification}. The distribution of all the features referred to in the same Section are in Figures~\ref{fig:biber_dist_part1} and~\ref{fig:biber_dist_part2}.  Figure~\ref{fig:ratio_plot_all} shows the ratio of aggregate feature usage across all LLMs to that in human-written text, over all features, as explained in Section~\ref{sssec:aggregate_overuse_underuse} (paragraph on visualization of usage ratios). Figure~\ref{fig:heat_map_all} plots the ratio of individual LLM feature usage relative to humans, across all features, as explained in Section~\ref{sssec:individual_overuse_underuse}. Figure~\ref{fig:dendrograms_per_category} visualizes the clustering of text sources for each individual genre as explained in Section~\ref{ssec:clustering}. Figure~\ref{fig:decoding_pca_all_categories} visualizes the effect of decoding strategies over all the genres, as explained in Section~\ref{ssec:decoding}.

\end{document}